\documentclass[conference]{IEEEtran}
\IEEEoverridecommandlockouts

\usepackage{cite}
\usepackage{amsmath,amssymb,amsfonts}
\usepackage{graphicx}
\usepackage{textcomp}
\usepackage{xcolor}
\usepackage{fancyhdr}

\def\BibTeX{{\rm B\kern-.05em{\sc i\kern-.025em b}\kern-.08em
    T\kern-.1667em\lower.7ex\hbox{E}\kern-.125emX}}
\usepackage{pifont}
\usepackage{amsmath,amsthm,amsfonts,amscd} 
\usepackage{tabularx}
\usepackage{tabu}
\usepackage{array}
\usepackage{color}
\usepackage{graphicx}
\usepackage{subfig} 
\usepackage{makecell}

\newcommand{\sect}[1]{Section~\ref{#1}}

\newcommand{\fig}[1]{Figure~\ref{#1}}
\newcommand{\tab}[1]{Table~\ref{#1}}

\newlength\dunder
\newcommand{\twound}{\rule{2\dunder}{0.4pt}}

\makeatletter
\newcommand{\thickhline}{%
    \noalign {\ifnum 0=`}\fi \hrule height 1.2pt
    \futurelet \reserved@a \@xhline
}
\makeatletter
\newcommand{\midhline}{%
    \noalign {\ifnum 0=`}\fi \hrule height 1pt
    \futurelet \reserved@a \@xhline
}

\usepackage{algorithm}
\usepackage[noend]{algpseudocode}

\floatevery{algorithm}{\setlength\hsize{4cm}}
\newcommand{\algo}[1]{Algorithm~\ref{#1}}

\algnewcommand{\LineComment}[1]{\State \(\triangleright\) #1}
\makeatletter
\renewcommand{\ALG@beginalgorithmic}{\footnotesize}
\makeatother

\algblockdefx{FORP}{ENDFP}[1]%
{\textbf{for all }#1 \textbf{do in parallel}}%
{\textbf{end for}}

\algblockdefx{FORB}{ENDFB}[1]%
{\textbf{for }#1 \textbf{do in backward}}%
{\textbf{end for}}

\begin{document}

\makeatletter
\def\ps@headings{
\let\@oddhead\@empty
\let\@evenhead\@empty
\def\@oddfoot{\@IEEEheaderstyle\hfil\thepage}%
\def\@evenfoot{\@IEEEheaderstyle\thepage\hfil\hbox{}}
}
\def\ps@IEEEtitlepagestyle{
\let\@oddhead\@empty
\let\@evenhead\@empty
\def\@oddfoot{\footnotesize Work in Process\hfill\thepage}%
\let\@evenfoot\@empty
}
\makeatother

\pagestyle{headings}
\setcounter{page}{1}
\thispagestyle{IEEEtitlepagestyle}

\author{
\IEEEauthorblockN{Sangkug Lym}                     
\IEEEauthorblockA{\textit{The University of Texas at Austin}\\                                               
sklym@utexas.edu}                                                                          
\and                                                                                    
\IEEEauthorblockN{Mattan Erez}                             
\IEEEauthorblockA{\textit{The University of Texas at Austin}\\                                               
mattan.erez@utexas.edu}                                                                          
}             

\title{FlexSA: Flexible Systolic Array Architecture for Efficient Pruned DNN Model Training}

\maketitle


\begin{abstract}


Modern deep learning models have high memory and computation cost. To make them fast and memory-cost efficient, structured model pruning is commonly used. We find that pruning a model using a common training accelerator with large systolic arrays is extremely performance-inefficient.

To make a systolic array efficient for pruning and training, we propose FlexSA, a flexible systolic array architecture. FlexSA dynamically reconfigures the systolic array structure and offers multiple sub-systolic operating modes, which are designed for energy- and memory bandwidth-efficient processing of tensors with different sizes and shapes. We also present a compilation heuristic for tiling matrix-multiplication-and-accumulation operations in a training workload to best utilize the resources of FlexSA. Based on our evaluation, FlexSA with the proposed compilation heuristic improves compute resource utilization of pruning and training modern CNN models by 37\% compared to a conventional training accelerator with a large systolic array. FlexSA also improves on-chip data reuse by 1.7X saving 28\% energy compared to naive systolic array splitting.

\end{abstract}


\section{introduction}
\label{sec:intro}

\emph{Neural network model pruning} is a commonly used technique to make deep learning inference fast and memory-efficient~\cite{han2015deep,han2015learning,li2016pruning,molchanov2016pruning}. It identifies non-critical parameters and removes them from a baseline model such that the pruned model has reduced computation and memory costs. This pruning procedure takes millions of iterations that repeat the process of pruning and retraining~\cite{he2017channel,hu2016network,molchanov2016pruning,he2018amc} or pruning while training~\cite{lym2019prunetrain,alvarez2017compression,gamboa2020campfire,NIPS2016_6504}. We find that commonly used deep learning training accelerators with large systolic arrays~\cite{jouppi2017datacenter,cyphers2018intel,tesla_ai_chip} are inefficient in processing these (intermediate) pruned models, which decrease compute unit utilization by 60\% in training and pruning modern convolutional neural network (CNN) models. This is true even for structured pruning, which removes parameters at the granularity of semantic groups (e.g., channels or network layers) such that the pruned models remain dense and do not require complex sparse data indexing that limits practical performance gains with arbitrary pruning~\cite{NIPS2016_6504,gray2017gpu}. To overcome this inefficiency of underutilized resources, we introduce {\bf{FlexSA}}, a flexible systolic array design that can reconfigure its structures at run-time to better fit the specific and dynamic workload of pruning DNN models and overcome the inefficiency we observe with standard systolic arrays.

Structured pruning removes parameters at the granularity of semantic groups (e.g., channels or network layers) and the pruned models remain dense, yet have fewer parameters. This pruning takes millions of iterations that repeat the process of pruning and retraining~\cite{he2017channel,hu2016network,molchanov2016pruning,he2018amc} or pruning while training~\cite{lym2019prunetrain,alvarez2017compression,gamboa2020campfire,NIPS2016_6504}. We find that commonly used deep learning training accelerators with large systolic arrays~\cite{jouppi2017datacenter,cyphers2018intel,tesla_ai_chip} are inefficient in processing these (intermediate) pruned models, which decrease compute unit utilization by 60\% in training and pruning modern convolutional neural network (CNN) models. To overcome this inefficiency, we introduce {\bf{FlexSA}}, a flexible systolic array design that can reconfigure its structures at run-time to better fit the specific and dynamic workload of pruning DNN models and overcome the inefficiency we observe with standard systolic arrays.

\begin{figure}[!t]
    \centering
    \includegraphics[width=0.48\textwidth]{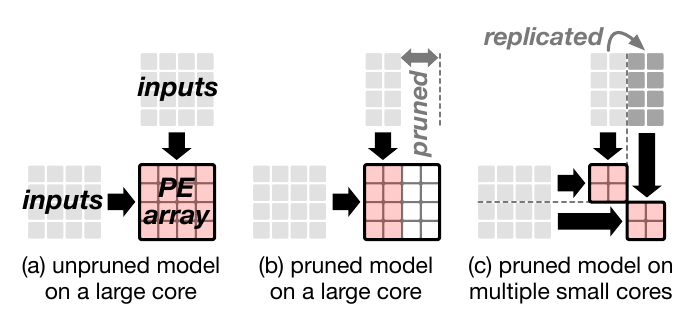}
    \caption{GEMM execution on systolic arrays with different sizes. The red PEs are utilized for GEMM execution and the white PEs are idle.}
    \label{fig:main}
\end{figure}

The execution inefficiency without FlexSA is mainly found in processing matrix multiplications and accumulations, which account for more than 95\% of the operations in modern deep learning models~\cite{he2016deep,szegedy2017inception,devlin2018bert,vaswani2017attention}. These matrix operations are commonly executed as general matrix multiplies (GEMMs)~\cite{lym2019delta} using high throughput training accelerators. For efficient acceleration of these GEMMs, many modern training accelerators adopt large systolic array cores that are (typically) a two-dimensional mesh of many simple and efficient processing elements (PEs)~\cite{jouppi2017datacenter,lym2018mini}. Because GEMM dimensions in model training are both large and multiples of the typical systolic array sizes, tiling and processing these GEMMs can fully utilize PEs on systolic arrays. This is shown in \fig{fig:main}{.a}, where the two arrays of gray boxes are input matrices for a tiled GEMM and each red box in the systolic array indicates an active PE.

However, the GEMM dimensions during structured pruning gradually and arbitrarily shrink. Thus, the tiles at the edges of an intermediate pruned GEMMs are not divisible by the size of a systolic array as shown in \fig{fig:main}{.b}. Processing these small GEMM tiles does not fully occupy the PEs (white boxes), which significantly decreases the performance. This problem is also known as \emph{tile quantization}, which seriously degrades performance in DNN training using GPUs~\cite{nvidia_dl}. Splitting a large systolic array into multiple small arrays can improve PE utilization (\fig{fig:main}{.c}). However, this naive core splitting requires broadcasting inputs to multiple cores. This input replication increases on-chip data traffic, decreasing the energy-inefficiency of training and pruning.

To achieve both the high on-chip reuse of using a large systolic array core and the high PE-utilization of using many-small cores,  FlexSA uses a core with four systolic arrays as the baseline. It adds input and output data paths between the four small cores and their control logic that incurs only 1\% area overhead. These design changes enable forming four different systolic array operating modes, which support inter-core collaborative and per-core individual systolic dataflows. These FlexSA modes improve input and output reuse by sharing them between cores, which makes FlexSA modes efficient in dealing with GEMM tiles with different sizes and shapes.

For efficient utilization of the flexible systolic array resources, we propose a compile-time GEMM tiling heuristic. The proposed heuristic prioritizes FlexSA modes that have high inter-core reuse. The FlexSA modes with limited inter-core reuse are chosen only when using them improves PE utilization. Our data shows that this compile-time technique uses inter-core systolic dataflows more than 89\% of the time without decreasing PE-utilization while pruning and training popular modern CNN models. The FlexSA compiler also handles loading inputs to input buffers then to registers and storing GEMM outputs to on-chip global buffers or DRAM at static time according to the type of FlexSA modes. Overall, the proposed FlexSA architecture increases PE utilization by 37\% compared to a large systolic array design and improves on-chip data reuse by 1.7X and energy efficiency by 28\% compared to a naive many-small-core design.

\medskip
To summarize, our main contributions are:
\begin{itemize}
\item
    We show that commonly used high throughput training accelerators with large systolic arrays are highly inefficient in processing structured model pruning due to reduced PE utilization.

\item
    We show that naively splitting a large core into many small cores decreases energy efficiency due to reduced on-chip reuse and performance of pruning because of increased memory bandwidth peaks.

\item
	We propose FlexSA, a flexible systolic array architecture that restructures its core structure to support multiple systolic array operating modes for efficient processing of GEMM tiles with different sizes and shapes.

\item
	We propose a compile-time GEMM tiling heuristic that tiles GEMMs into parts that best utilize FlexSA resources. FlexSA with our proposed GEMM tiling heuristic achieves both high data reuse of using a large core and high PE utilization of using multiple small cores.


\end{itemize}

\section{background}
\label{sec:background}

In this section, we briefly discuss the background on deep learning applications and training accelerator architectures needed to understand the challenges we focus and our proposed ideas for them.

\subsection{Network Model Training and Pruning}
\label{subsec:background_pruning}
Deep learning applications consist of many layers that have millions of weights (or learning parameters) to learn complex features from a large training dataset. Training needs forward and back-propagating the training dataset through network layers repetitively, which involves trillions of FLOPs (floating-point operations) and memory accesses. Most of these FLOPs are matrix multiplications and accumulations and they are typically processed as GEMMs for simple and efficient processing on high-throughput training accelerators. Training generally uses large mini-batch sizes (a set of input samples propagated at a time per training iteration) thus the GEMMs in training have large data parallelism in all dimensions~\cite{keskar2016large,neyshabur2017exploring}.

\medskip
\noindent\textbf{Model Pruning Methods.}
Model pruning has been studied primarily for CNNs (convolutional neural networks), to make their models more compact and their inference fast and energy-efficient. Pruning methods compress a CNN model by removing non-critical (or small-valued) weights by identifying such weights during training \cite{alvarez2017compression,NIPS2016_6504} or after training using a pre-trained baseline model~\cite{he2017channel,hu2016network,molchanov2016pruning,he2018amc}. They gradually remove non-critical weights by repeating the process of pruning and re-training to minimize its impact on accuracy. Recent prior work prunes weights during training to save not only the cost for inference but also for training~\cite{lym2019prunetrain,gamboa2020campfire}. They regularize weights to small values then prune the weights whose absolute values are under a small threshold. For both approaches, pruning happens gradually thus training accelerators have to deal with both the baseline model and (intermediate) pruned models.

These pruning algorithms can be unstructured or structured. Unstructured (or individual-weight) pruning can maximize model-size reduction but requires fine-grained indexing with irregular data access patterns. Such accesses and extra index operations lead to poor performance on deep learning accelerators with vector or matrix computing units despite the reduced number of weights and FLOPs~\cite{han2017ese,yu2017scalpel,anwar2017structured}. 

On the other hand, structured-pruning algorithms remove or reduce fine-grained indexing and better match the needs of hardware and thus effectively realize performance gains. \emph{Channel pruning} is the most commonly used structured pruning technique for CNNs~\cite{he2017channel,NIPS2016_6504,he2018amc,howard2017mobilenets,lym2019prunetrain,zhuang2018discrimination,gao2018dynamic}. It removes weights at the granularity of channels. This reduces one of three GEMM dimensions (width, height, and depth) of each convolution layer reducing its data parallelism.

\subsection{Systolic Arrays}
\label{subsec:background_systolic}

\begin{figure}[!t]
    \centering
    \includegraphics[width=0.48\textwidth]{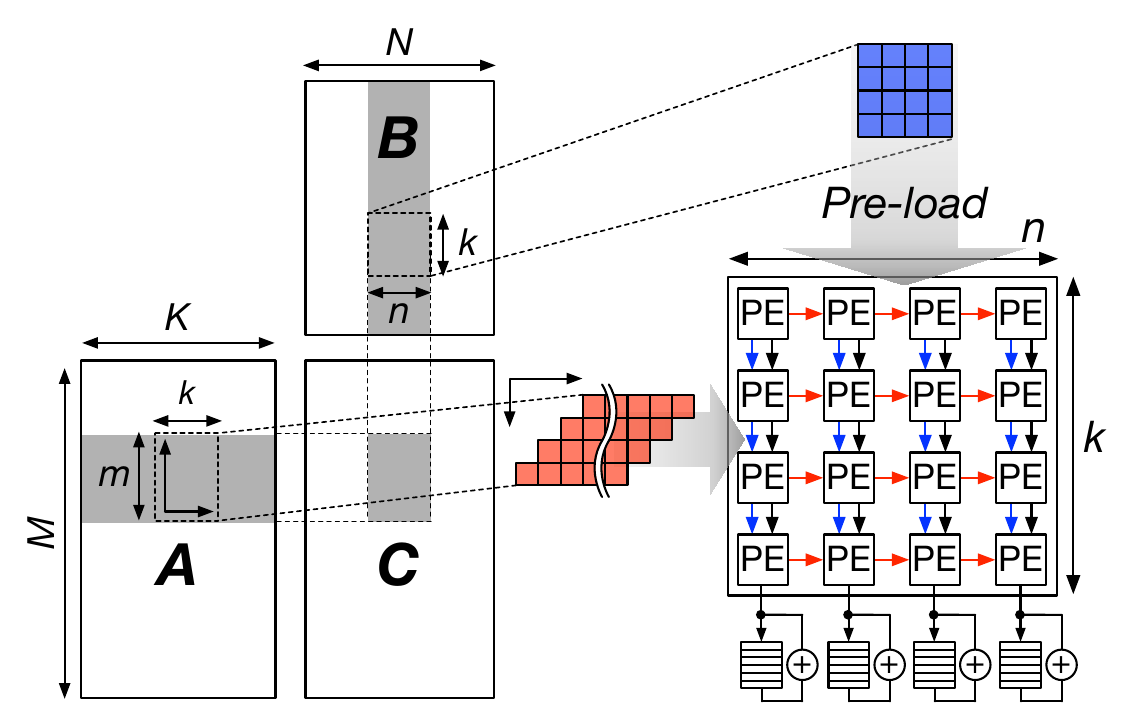}
    \caption{GEMM tiling using the size of systolic array and \emph{systolic wave} execution using input-stationary systolic dataflow.}
    \label{fig:systolic_arch}
\end{figure}

Training accelerators are designed to minimize training time thus requires high compute throughput. To efficiently process GEMMs in training workloads using a large number of PEs while avoiding redundant memory accesses, many modern training accelerators (e.g., Google's TPU~\cite{jouppi2017datacenter}, Intel's Nervana NNP~\cite{cyphers2018intel}, and Tesla's AI chip~\cite{tesla_ai_chip}) adopt systolic arrays. A systolic array (\fig{fig:systolic_arch}) is a two-dimensional mesh of many simple and efficient PEs. At each cycle of kernel execution, each PE applies the same computation to its inputs and then passes the computed result or its unmodified inputs to one or more of its neighbors. All PEs communicate only with adjacent PEs such that there is minimal data movement and high computational concurrency~\cite{kung1982systolic}. A GEMM is tiled such that each tile is processed fully within the systolic array. Ideally the tile is sized to exactly fit the array and fully utilize its PEs. Tiles are scheduled according to a fixed systolic dataflow. Input- and output-stationary systolic dataflows are the most commonly used ones and achieve high in-core input and output reuse, respectively~\cite{samajdar2018scale}.

Computation of systolic dataflows consists of pipelining inputs from the top and left edges of the array (in our example) and obtaining results at each PE or at the bottom of the array depending on the dataflow type. \fig{fig:systolic_arch} shows the operation of input-stationary dataflow, where one of the two input matrices for a GEMM tile is pre-loaded at each PE then the other input matrix is shifted horizontally. Executing a GEMM tile is further blocked by the local input buffer size, processing \emph{m} at a time. We call this blocked systolic computation batch a {\bf{systolic wave}}.

\section{Architectural Challenges of \\Model Pruning}
\label{sec:flexsa_motivation}
\index{flexsa_motivation}

Channel pruning is done by retraining a pre-trained dense model or pruning while training from scratch. In both cases, training accelerators have to deal with both dense and (intermediate) pruned models. Unlike unpruned models that typically have a regular number of channels, such as 64, 128, or 256 (powers of two)~\cite{krizhevsky2012imagenet,he2017mask,redmon2016you}, their channel-pruned versions have an irregularly reduced number of channels (e.g., 3, 71). The GEMMs converted from such channel-pruned convolution and FC (fully-connected) layers also have severely reduced sizes. Tiling such GEMMs to large systolic array cores can be inefficient: many GEMM tiles have sizes smaller than the size of a large systolic array (e.g., 128$\times$128). Thus processing these tiles does not fully utilize the PEs in the systolic array. This causes severe compute resource underutilization and limits training time saving even with large FLOPs reduction by continuous pruning.


\begin{figure}[t!]
    \centering                                                                            
    \subfloat[Pruning strength = Low]{
        \includegraphics[width=0.44\textwidth]{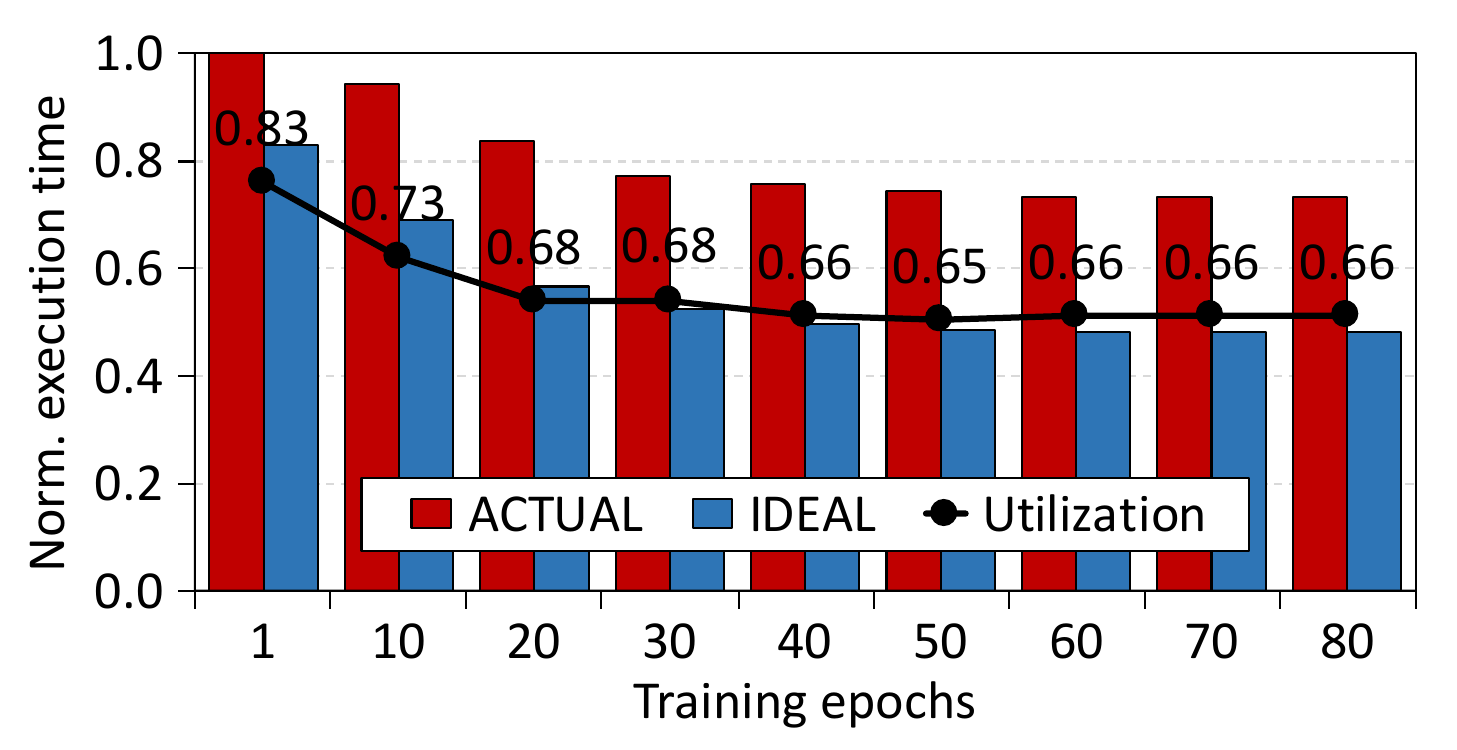}
        \label{fig:flexsa_pe_util_ideal}
    }\\
    \vspace*{-2.0mm}
    \subfloat[Pruning strength = High]{
        \includegraphics[width=0.44\textwidth]{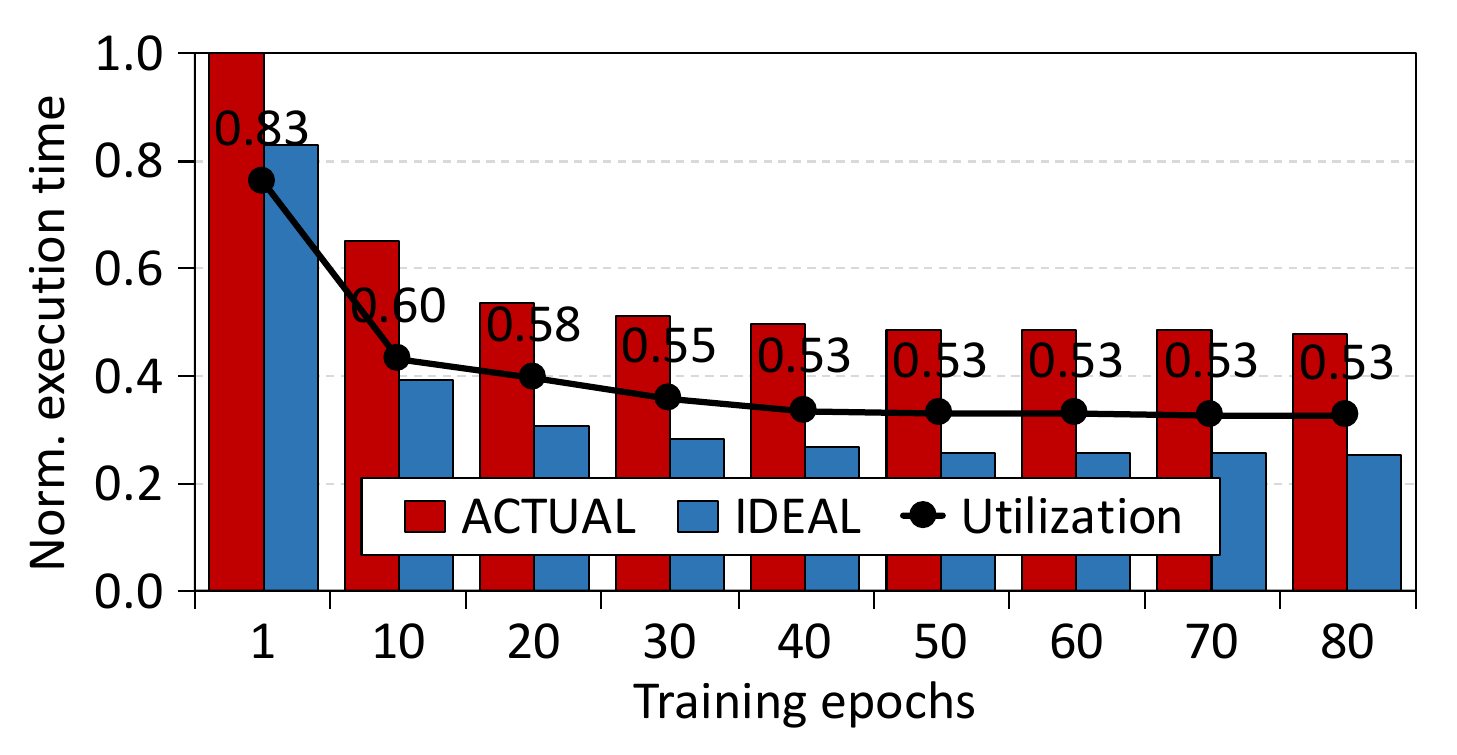}
        \label{fig:flexsa_pe_util_hbm}
    }
    \caption{Execution time of channel-pruned ResNet50 normalized to the execution time of the unpruned baseline.} 
    \label{fig:pe_underutilization}
\end{figure}


\begin{figure*}[h]
\centering
\begin{minipage}{0.56\linewidth}
    \includegraphics[width=\textwidth]{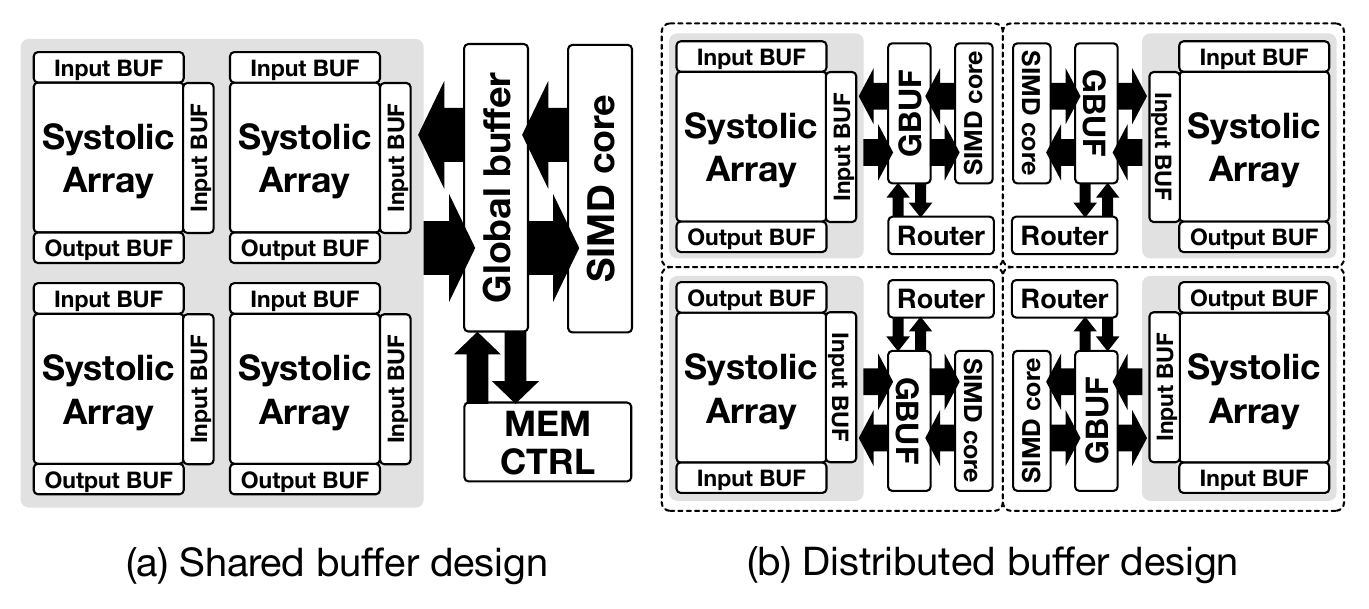}
    \vspace*{-4mm}
    \caption{Different global buffer designs.}
    \label{fig:buffer_design}
\end{minipage}
\begin{minipage}{0.43\linewidth}
    \includegraphics[width=\textwidth]{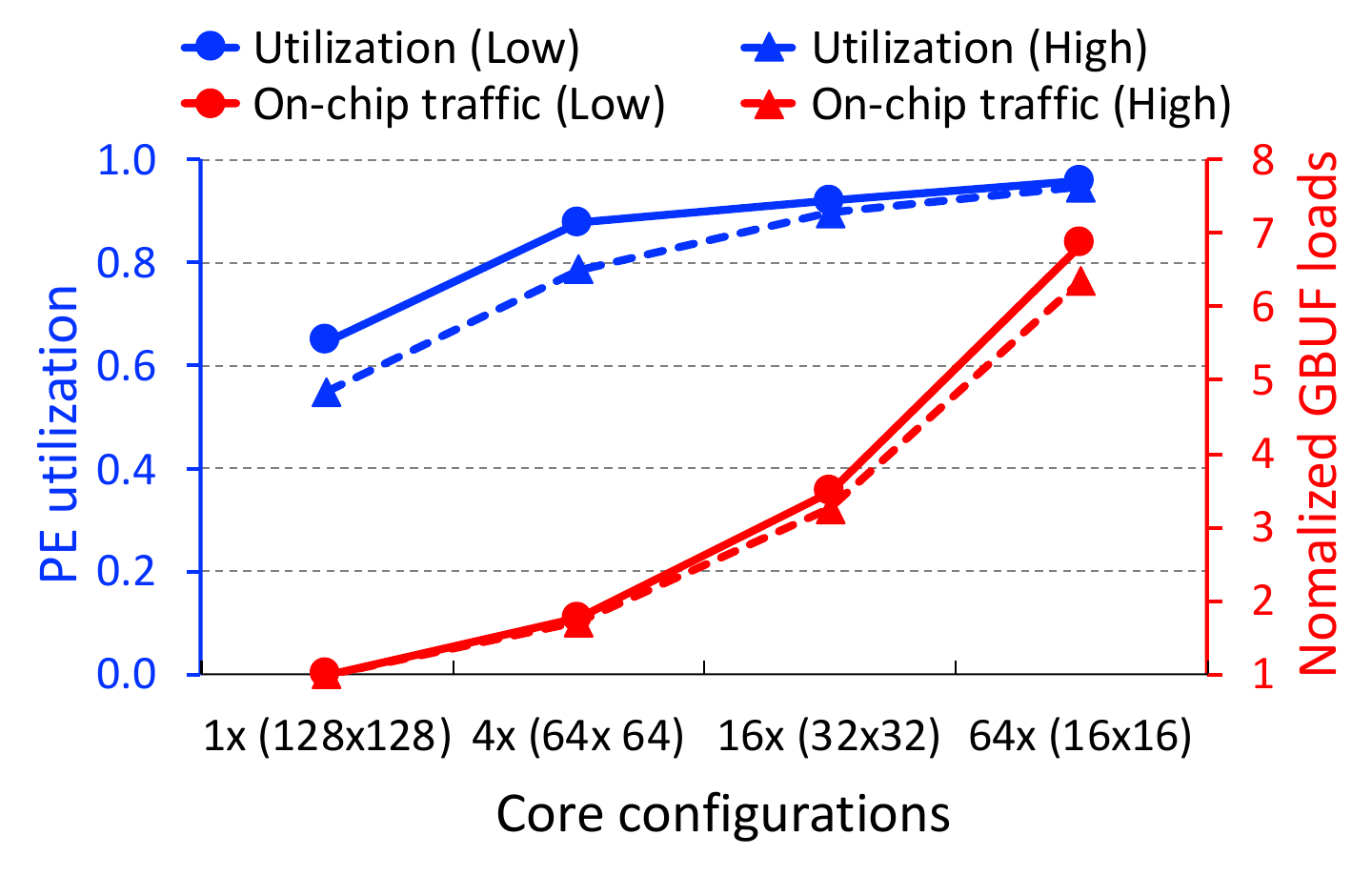}
    \vspace*{-5mm}
    \caption{Impact of core sizing to PE utilization and on-chip data traffic.}
    \label{fig:core_sweep}
\end{minipage}

\vspace*{-2mm}
\end{figure*}

To show the low efficiency of a large systolic array for training pruned CNN models, we prune ResNet50~\cite{he2016deep}, the most commonly used CNN model for image classification, while training. We use the {\bf{WaveCore}} training accelerator, discussed by Lym et al.~\cite{lym2018mini}, which executes GEMM tiles using 128$\times$128 systolic arrays using the input-stationary dataflow, similar to Google's TPU v3~\cite{tpu_v3}. We use {\bf{PruneTrain}}, which regularizes parameters in each channel to zero as a mechanism to prune channels during training~\cite{lym2019prunetrain}. \fig{fig:pe_underutilization} shows the execution time per training iteration over the process of pruning-while-training using WaveCore. Each bar shows only the sum of convolution and FC layer execution time considering that they account for \(>\)98\% of the FLOPs in a training iteration. The training iteration time is normalized to that of the baseline before pruning. In this experiment, we use a pruning interval of 10 epochs and the two figures show the results of using different pruning strengths as introduced in PruneTrain: low pruning strength that removes relatively few channels with only a small accuracy loss (top), and high pruning strength that removes more channels with a larger accuracy loss (bottom). We evaluate this using an instruction-level simulator that we develop.


The figures show two bar graphs: the blue bars (\emph{IDEAL}) indicate the execution time with the assumption of 100\% PE utilization and the red bars (\emph{ACTUAL}) show the increased execution time caused by PE underutilization. All bars are normalized to the execution time of the unpruned baseline (the left-most red bar in each figure). The black line in each figure indicates the PE utilization at each pruning interval. The PE underutilization is solely caused by the size mismatch between GEMM tiles and the systolic array and is estimated with ideal (infinite) memory BW. 

As the blue bars show, PruneTrain gradually reduces the FLOPs of the unpruned baseline model to 48\% and 25\% when using the low and high pruning strengths, respectively. However, due to the large systolic array size and reduced GEMM dimensions in many convolution layers, PEs become highly underutilized as pruning proceeds and exhibit overall PE utilization of only 69\% and 58\% for the low and high pruning strengths, respectively. Also, even the baseline model shows only 83\% PE utilization and this is because some early convolution layers have a small number of channels.

\section{Naive GEMM Core Splitting}
\label{sec:flexsa_core_splitting}

\medskip\noindent{\textbf{Diminishing PE Utilization and Increasing Input Load Cost.}}
The PE underutilization caused by small GEMM dimensions can be mitigated by splitting a large systolic array core into multiple small cores. GEMMs can be tiled into smaller parts then executed with higher PE occupation on the smaller cores. However, this many-small-core design can be inefficient in processing large GEMM tiles, which still account for most GEMM tiles in both pruned and unpruned CNN layers. First, small cores have low in-core input reuse compared to a large core, which increases input traffic from memory or the global buffer. Second, having more cores increases the area complexity of wires, data path switches, and SRAM buffer control and decoding logic.

To study the relationship between the core size and PE utilization, we conduct an experiment that estimates overall PE utilization and the volume of on-chip data traffic while pruning ResNet50 during training with PruneTrain. The baseline accelerator design used in this experiment has a group of systolic array cores sharing one global buffer (GBUF) (\fig{fig:buffer_design}{.a}). The GBUF is used to block GEMM inputs and share them among cores. The evaluation is done with different core configurations, which are the points on the X-axis of \fig{fig:core_sweep} indicating $\#$cores$\times$core size. This baseline design also has a pair of local buffers (LBUFs) in each core for input double buffering to hide input load latency. The blue lines in \fig{fig:core_sweep} show PE utilizations and the red lines show the average data traffic volume between the GBUF to LBUFs when using low (solid lines) and high (dotted lines) pruning strengths. Both PE utilization and data traffic volume are averaged across training iterations over the whole training run, thus they encapsulate the results of the baseline unpruned model and many different intermediate pruned models. We use ideal memory BW to show the PE underutilization caused solely by the size mismatch between GEMM tiles and the core.

Splitting one 128$\times$128 core into four 64$\times$64 improves PE utilization by as much as 23\% but it also increases input load traffic by 1.7$\times$ due to reduced input reuse. Further splitting the cores yields diminishing PE utilization improvement but continues to increase data traffic. Using 16$\times$ (32$\times$32) and 64$\times$ (16$\times$16) cores improves PE utilization by only 8\% and 4\% but increases input load traffic by 3.4$\times$ and 6.6$\times$. Also, 32$\times$32 and 16$\times$16 cores show almost similar PE utilization for the two ResNet50 pruned using different pruning strengths. This indicates that small cores do not necessarily improve the PE utilization of the more aggressively pruned CNN models. \emph{Overall, this result shows that reducing the core size to less than 32$\times$32 is not cost-efficient.}

\medskip\noindent{\textbf{High Area Overhead of the Many-Small-Core Design.}}
The increased data traffic of small cores requires higher on-chip data BW: when the core count increases by 4$\times$, the peak on-chip data BW required increases by 2$\times$. The impact of increasing input BW is different depending on the accelerator design. \fig{fig:buffer_design} shows two potential buffer designs. In the baseline shared buffer design (design (a)), cores share a GBUF. The distributed buffer design (design (b)), there is a dedicated GBUF for each core. A shared buffer needs additional data paths from GBUF to the LBUF for each core, along with path switches. On the other hand, a distributed buffer design has lower data path-associated area overhead but instead increases area for splitting LBUFs and GBUFs (duplicating decoding and data repeating logic). Furthermore, in the distributed buffer design, inputs should be carefully allocated to cores to avoid input replication. Although shared data can be split across cores, this leads to data transfers over potentially low-BW inter-core channels. A more general accelerator architecture mixes these two designs by creating multiple groups of cores such that cores in each group share a GBUF~\cite{lindholm2008nvidia,gao2017tetris}.

\begin{figure}[!t]
    \centering
    \includegraphics[width=0.47\textwidth]{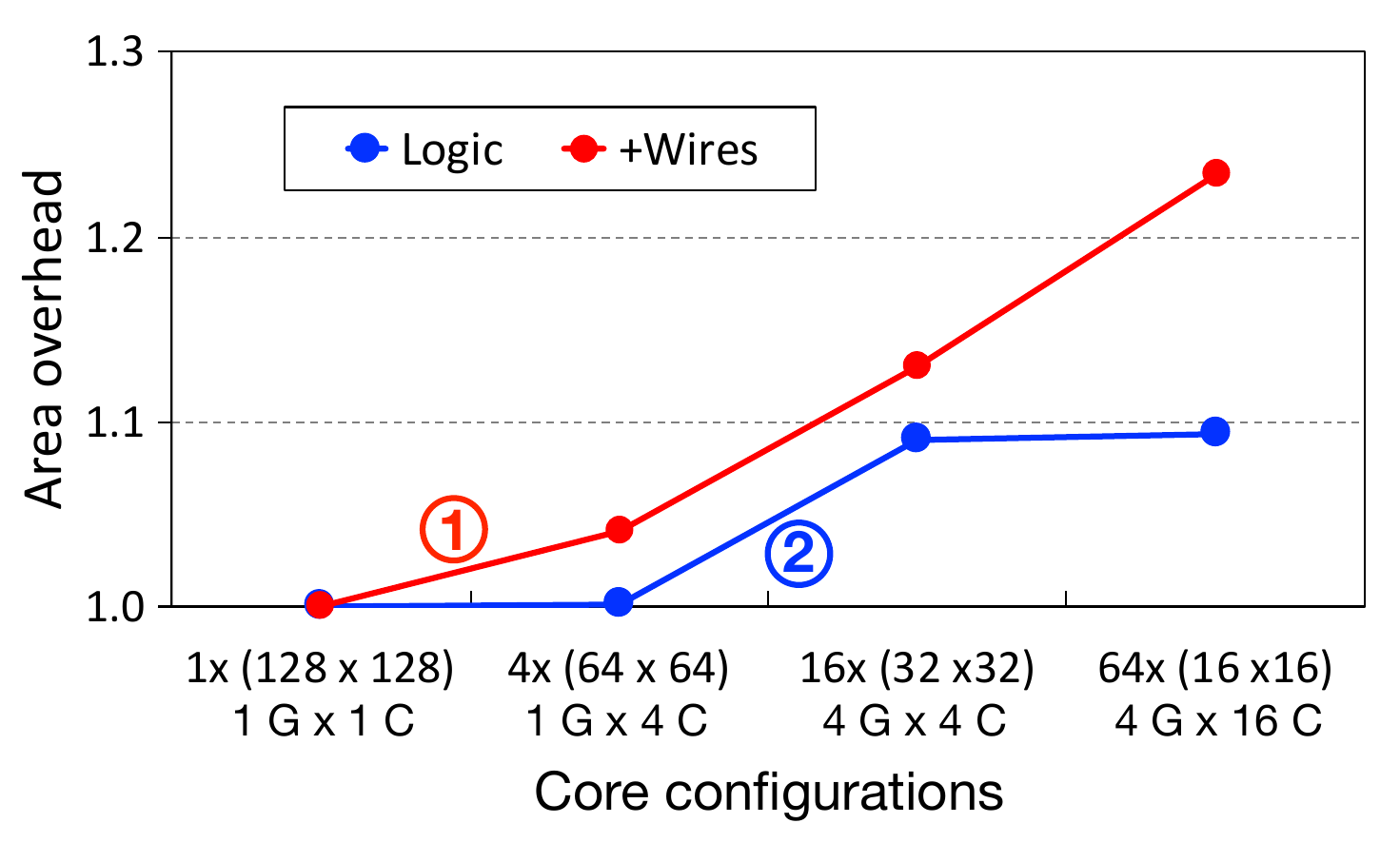}
    \caption{Area overhead of splitting a large core to multiple small cores. Overheads are normalized to 1$\times$ (128$\times$128).}
    \label{fig:area}
\end{figure}

We compare the area overhead of different core configurations that have different core sizes and buffer designs (\fig{fig:area}). For simple estimation, this experiment considers only the area of PEs, SRAM buffers, and data paths. We use CACTI 7.0~\cite{chen2012cacti} to estimate the area overhead of buffers and use the module size of a mixed-precision multiplier and adder unit~\cite{zhang2018efficient} to estimate the area of the systolic array. When the number of cores becomes greater than four, we group the cores such that they share a GBUF and this configuration is shown in the second row of the X-axis (G and C indicate groups and cores, respectively). The blue line graph in \fig{fig:area} shows the area overhead of only the additional logic for splitting GBUFs and LBUFs. The red line indicates the area overhead of increased data paths.

The overhead of data paths is estimated conservatively assuming the wires do not overlap with logic (so the actual overhead should be smaller). The increase in chip width and height from additional data paths is estimated by distributing the wires to 5 metal layers using a wire pitch of 0.22um, similar to the method used by DaDianNao~\cite{chen2014dadiannao}. The overheads of different core configurations are normalized to the area of a single 128$\times$128 core.

Splitting a single core to 4 (64$\times$64) cores has a relatively small area overhead of 4\%, and this (\textcolor{red}{\ding{182}}) mainly comes from doubling the data paths by having cores share a GBUF. Further splitting them to 16 (32$\times$32) cores increases area overhead to as much as 13\%, where the overhead for dividing a GBUF to four parts is added (\textcolor{blue}{\ding{183}}). Finally, 64 (16$\times$16) cores have a 23\% area overhead. The additional area overhead comes from increasing the number of cores sharing a GBUF in each group. \emph{Overall, this experiment shows that splitting a large core is not scalable. Especially, splitting a core into \(\geq\)16$\times$ small cores is not an area-efficient design option.}

In summary, naively splitting a large core into many small cores improves PE utilization but its gain diminishes as the splitting factor increases. In addition, the many-small-core design increases total traffic from the GBUF due to the reduced in-core input reuse and also requires high area overhead. To better balance these tradeoffs, \emph{a new systolic array core architecture is needed to maintain high input reuse as using a large core and achieve high mapping flexibility as when using multiple small cores.}



\section{Flexible Systolic Array}
\label{sec:flexsa_architecture}

In this section, we introduce a systolic array architecture that achieves both high PE utilization and low on-chip traffic in processing (partially) pruned CNN models.

\subsection{FlexSA Core Architecture}
\label{subsec:sub-array op}

To map small GEMM tiles to cores with high PE utilization, splitting a large core into smaller cores is unavoidable. However, processing a large GEMM tile using multiple small cores is inefficient as this increases on-chip data traffic between GBUFs and cores. Especially, this becomes a critical problem as GBUF access costs grows with the size of the systolic array. To take advantage of the benefits of both a large core and multiple small cores, our idea is to use a group of small cores that collaborate when processing a large GEMM tile and work independently when processing small GEMM tiles. To this end, we propose {\bf{FlexSA}}, a flexible systolic array architecture that meets the need for both high input reuse and PE utilization.

\begin{figure}[!t]
    \centering
    \includegraphics[width=0.38\textwidth]{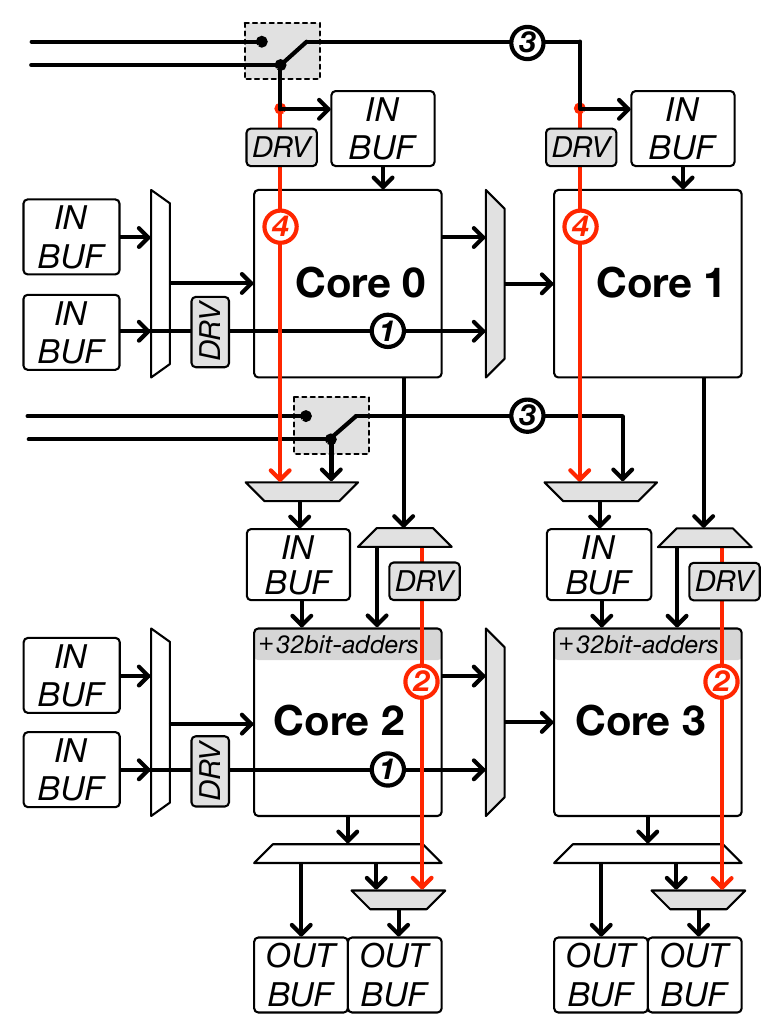}
    \vspace*{-1.0mm}
    \caption{FlexSA architecture.}
    \label{fig:core_structure}
    \vspace*{-2.0mm}
\end{figure}

\fig{fig:core_structure} shows the logical structure of FlexSA. FlexSA is based on four systolic array cores sharing a GBUF, as illustrated in \fig{fig:buffer_design}{.a}. However, unlike the baseline four-core design, where each core always operates independently, FlexSA can reconfigure its operation mode such that the four cores work independently or collaboratively. FlexSA provides four different operating modes and they are supported using additional data paths and path switches (red data paths in \fig{fig:core_structure}).

\begin{figure}[!t]
    \centering
    \includegraphics[width=0.48\textwidth]{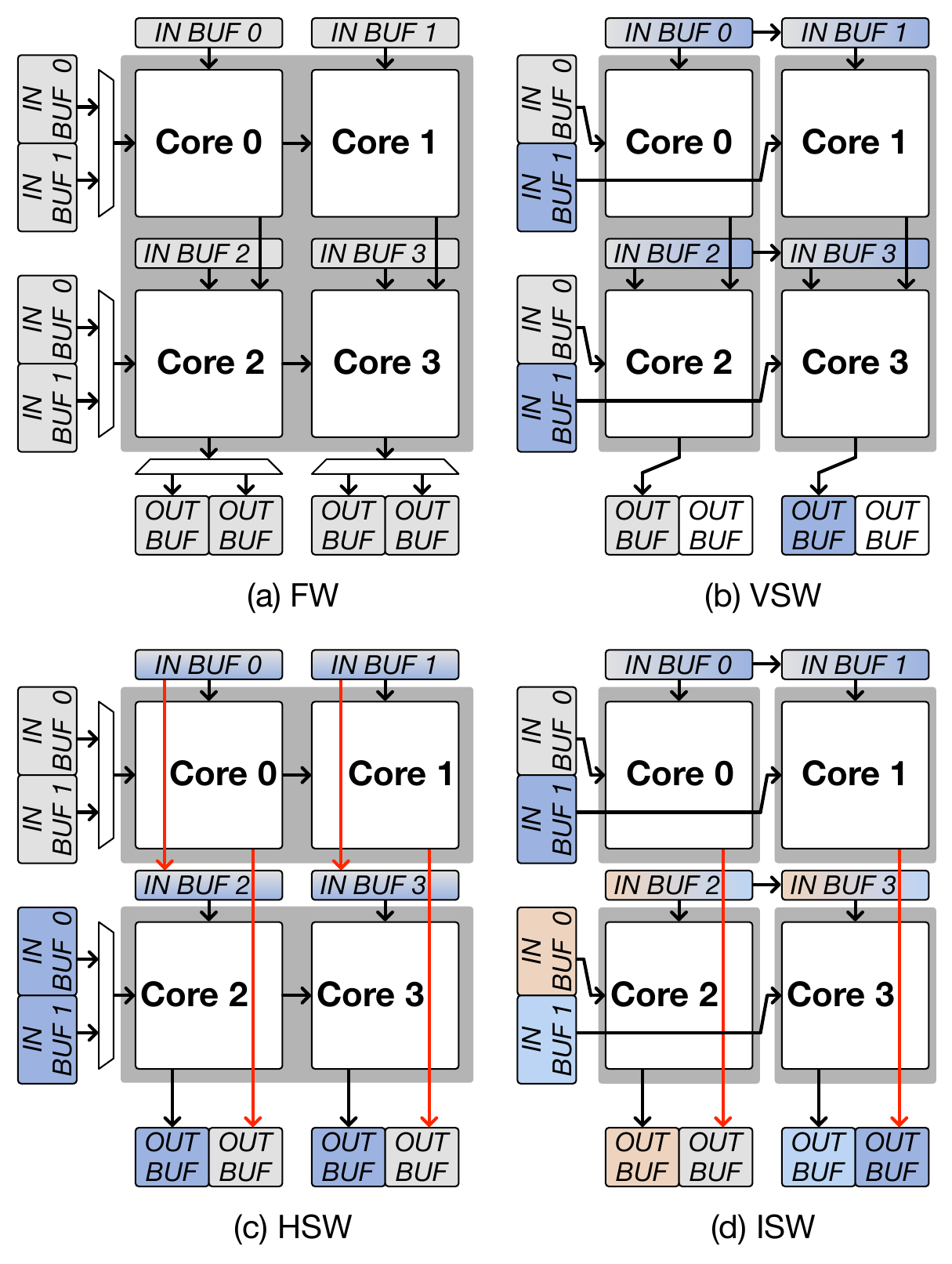}
    \caption{Four different systolic sub-array operations supported by FlexSA and the micro-architecture settings for each mode. Buffers with different colors hold inputs and outputs of different waves. The color gradients indicate that inputs in the buffers are shared between two waves.}
    \label{fig:flexsa_ops}
    \vspace*{-2.0mm}
\end{figure}

\medskip\noindent{\textbf{Sub-Array Operations.}}
The four different systolic operating modes are illustrated in \fig{fig:flexsa_ops}. {\bf{FW}} (full wave) uses the four cores as a single systolic array by sharing inputs and passing partially accumulated outputs between cores (\fig{fig:flexsa_ops}{.a}). First, the inputs are vertically shifted from the LBUF on top of each core. These inputs are all unique and they are pre-loaded to each PE for the input-stationary systolic dataflow. Compared to a single large systolic array, this saves half the input shifts due to the reduced core height. Next, both cores 0 and 2 pass the inputs shifted from the left LBUFs to core 1 and 3 for reuse in multiplication and accumulation operations (with the stationary inputs pre-loaded in core 1 and 3). At the same time, cores 0 and 1 pass their outputs (partial sums) directly to cores 2 and 3 for output reuse. Because the reuse of both vertically and horizontally shifted inputs is the same as the large core, FW has half the on-chip traffic compared to a naive independent four-core design.

FlexSA supports two systolic sub-array operations that use two pairs of cores independently. First, {\bf{VSW}} (vertical sub-wave) forms two vertical systolic sub-arrays by paring the two cores in each column (cores 0 and 2 for one sub-array and cores 1 and 3 for the other as shown in \fig{fig:flexsa_ops}{.b}). This mode is designed for efficient processing of skinny GEMM tiles in parallel whose tile width is smaller than or equal to the width of one small core. VSW starts by pre-loading the same inputs to each of the two vertical sub-arrays. To reduce the cost of sending identical inputs from GBUF to the LBUFs of both sub-arrays, we construct switchable data paths between the LBUFs of cores 0/2 and 1/3 for local broadcast \ding{184}. After pre-loading the stationary inputs, the other inputs are horizontally shifted to each sub-array. To provide different inputs to each sub-array in parallel, VSW uses the additional horizontal data paths \ding{182}. VSW uses only half of the output buffers, as shown in \fig{fig:flexsa_ops}{.b}. Therefore, it is possible to interleave two VSWs that use the same stationary inputs but different horizontally shifted inputs. Overall, VSW improves PE utilization by executing two skinny GEMM tiles in parallel and also improves stationary input reuse by 2$\times$ through local broadcasting.

Second, {\bf{HSW}} (horizontal sub-wave) constructs two horizontal systolic sub-arrays by paring the two cores in each row (cores 0/1 for one sub-array and cores 2/3 for the other as shown in \fig{fig:flexsa_ops}{.c}). HSW is designed for efficient mapping of fat GEMM tiles whose accumulation depth is smaller than or equal to the height of one small core. HSW first pre-loads the same stationary inputs to each horizontal sub-arrays. Then, the other inputs are shifted from the left LBUFs through both paired cores in each row. Similar to the local input sharing used in VSW, we construct a direct data path between the input buffers of cores 0/1 and cores 2/3 to avoid duplicated input transfers from the GBUF \textcolor{red}{\ding{185}}. The outputs from cores 0 and 1 are directly stored and accumulated at the output buffers at the bottom of cores 2 and 3 \textcolor{red}{\ding{183}}. Overall, HSW improves PE utilization by processing two fat GEMM tiles in parallel and doubles the reuse of horizontally shifted data compared to using small cores individually.

Lastly, FlexSA supports full independent core operation by providing unique inputs to each core as shown in \fig{fig:flexsa_ops}{.d}, which we call {\bf{ISW}} (independent sub-wave). This mode helps maintain high PE utilization when both GEMM tile width and accumulation depth are small. ISW provides independent inputs horizontally to cores 1 and 3 using the additional data path \ding{182} and the outputs of core 1 and 2 are sent to and accumulated at the output buffers using the added vertical data path \textcolor{red}{\ding{183}}. Compared to the independent core design, ISW exhibits lower input loading cost because FelxSA locally broadcasts the stationary inputs between cores. This saving is possible because the FlexSA compiler schedules waves sharing one of the two inputs in parallel rather than processing waves with all different inputs (we will discuss this in \sect{sec:gemm_inst_selection}. However, since the input reuse of ISW is the lowest among all modes, other sub-array modes should be prioritized over ISW for cost-efficient execution.

\subsection{Area Overhead Estimation}
\label{subsec:area overhead}
FlexSA requires additional area for logic and data path wires and switches. We estimate area in 32\(nm\) technology and we compare the overhead relative to the naive four-core design. Again, we conservatively estimate the area overhead of additional data paths assuming they do not overlap with logic. First, the addition of data path switches (1:2 MUXs) that select inputs and partial sums increase logic area by only 0.03\(mm^2\). Second, FlexSA requires each PE at the top row of cores 2 and 3 to support a mixed-precision FMA (fused multiplier and adder) instead of just a 16-bit multiplier to accumulate the partial sums shifted from cores 0 and 1. This change increases the area by 0.32\(mm^2\). Third, the repeaters to drive signals over a core add 0.25\(mm^2\), where we use fanout of 32. The vertical wires connecting the outputs of cores 0 and 1 to the output buffers \textcolor{red}{\ding{183}} expand the width of the core by 0.09\(mm\). However, the other wires do not affect die size because they are effectively canceled by the wiring overhead of connecting GBUFs and the LBUFs in the baseline four-core design. 

Overall, FlexSA increases the die area over the naive four-core design by only 1\%. This area estimate is conservative and placing the newly added vertical wires over PE array (as illustrated) can effectively hide the wiring area overhead. This is feasible because PE arrays use only a few low-level metal wires for local routing.


\medskip\noindent{\textbf{Design Scalability.}}
It is possible to form a FlexSA unit using more than four cores. However, we present only the four-core design for three reasons. First, the performance benefit diminishes as we further split the cores to a larger number of smaller cores as shown in \fig{fig:core_sweep}. Second, grouping more than four cores incurs significantly higher area overhead. This area is for the larger number of wires, data switches, and repeaters required for constructing many independent sub-arrays Instead, we propose to construct multiple FlexSA units with four cores, which incurs no additional area overhead, making it scalable. We use this multi-FlexSA design to evaluate model pruning in \sect{sec:eval}.

\section{FlexSA Compilation and GEMM Tiling}
\label{sec:gemm_inst_selection}

FlexSA mode selection is crucial for efficient GEMM processing. Modes should be selected to achieve the highest PE utilization with minimum on-chip data traffic. Mode selection is coupled with GEMM tiling because different modes are optimal for GEMM tiles with different shapes.

In this section, we introduce a compile-time GEMM tiling heuristics that tile a GEMM into {\emph{systolic waves}} to best utilize FlexSA resources such that it achieves both high performance and energy efficiency. Here, a wave the GEMM execution granularity using the systolic array defined in \sect{sec:background}. The systolic waves are mapped to systolic array cores and executed using different FlexSA modes. Different modes require different micro-controls, such as data path selection, input shifting, output shifting, and partial sum accumulation. For efficient communication between software and the FlexSA micro-architecture, we introduce a set of instructions that define the micro-controls needed for each FlexSA mode and that handle data transfers between on-chip buffers.

\subsection{GEMM Tiling Heuristics}
\label{subsec:gemm_heuristics}
The proposed compile-time GEMM tiling heuristics choose waves that are executed on FlexSA with the highest PE utilization and in-core data reuse. In other words, systolic waves that use the FlexSA modes with greater input reuse are prioritized and the waves using the FlexSA modes with lower input reuse are chosen only when using them improves PE utilization (i.e., FW$>$HSW=VSW$>$ISW).

\begin{figure}[!t]
    \centering
    \includegraphics[width=0.44\textwidth]{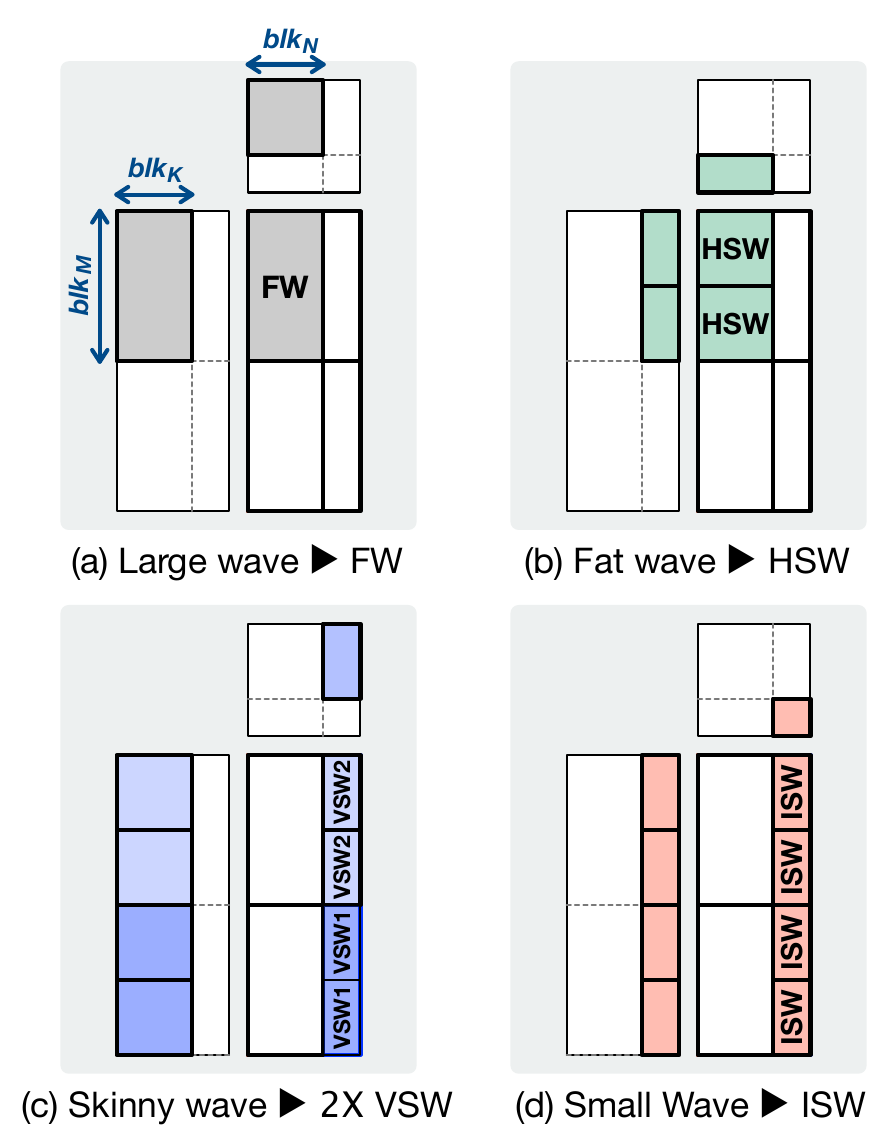}
    \vspace*{-1.0mm}
    \caption{FlexSA operating mode selection examples.}
    \label{fig:flexsa_insts}
\end{figure}

\medskip\noindent{\textbf{GEMM Tiling Conditions for FW.}}
The proposed heuristics tile a large GEMM into multiple waves to process them on the cores of FlexSA. The tiling factors of all GEMM dimensions (\(blk_M\), \(blk_N\), and \(blk_K\) in \fig{fig:flexsa_insts}{.a}) are chosen to match the size of a full FlexSA (all four cores of FlexSA) to maximize reuse by utilizing the FW mode. The sizes of \(blk_N\) and \(blk_K\) are equal to the width and height of a full FlexSA core, both of which are 128 in our baseline design. The size of \(blk_M\) is the size of the LBUF for non-stationary inputs (LBUFs on the left side of FlexSA in \fig{fig:core_structure}) divided by the height of a full FlexSA core. 

Because GEMM dimensions are not always a multiple of the ideal tile size, the execution of some tiles with FW would degrade utilization and performance. Such tiles (typically just the edge tiles of a large GEMM) are executed with the other FlexSA modes as shown in \fig{fig:flexsa_insts} and discussed below. 

\medskip\noindent{\textbf{GEMM Tiling Conditions for HSW.}}
When the \(blk_K\) size of a systolic wave is small, using FW under-utilizes PEs and slows down the GEMM execution. Thus, when the size of \(blk_K\) is smaller than or equal to the height of one FlexSA core (a sub-core of FlexSA), the proposed heuristics execute the systolic wave using HSW as shown in \fig{fig:flexsa_insts}{.b}. By executing a single fat wave as two independent waves using the two horizontal sub-systolic arrays, HSW increases PE utilization by 2X.

\medskip\noindent{\textbf{GEMM Tiling Conditions for VSW.}}
Tiles at the GEMM edges can have small block sizes in the \(blk_N\) dimension and executing the systolic waves of these skinny tiles using FW also under-utilizes PEs. Therefore, when the \(blk_N\) is smaller than or equal to the width of one FlexSA core, the heuristics execute the waves of the skinny GEMM tiles using VSW as shown in \fig{fig:flexsa_insts}{.c}. This improves PE utilization by executing the waves of two small GEMM tiles in parallel using the two vertical sub-systolic arrays. In particular, this skinny GEMM tile is executed using two VSW operations (VSW1 and VSW2 in \fig{fig:flexsa_insts}{.c}) that can be interleaved, accumulating their results using half of the output buffers. Because the two VSW operations share the stationary inputs, this achieves additional 2X stationary input reuse.

\medskip\noindent{\textbf{GEMM Tiling Conditions for ISW.}}
When both \(blk_N\) and \(blk_K\) are smaller than the width and the height of one FlexSA core, respectively, the proposed heuristics execute the systolic waves using ISW. ISW executes a GEMM tile using four small independent waves, which operate in parallel to achieve high PE utilization (\fig{fig:flexsa_insts}{.d}). When a pair of VSWs precedes ISW like the case of \fig{fig:flexsa_insts}{.c} followed by \fig{fig:flexsa_insts}{.d}, the four small systolic waves executed by ISW accumulate their results with the partial sums pre-stored in the output buffers. 

\subsection{ISA Support and Compilation for GEMM Processing}
To execute the systolic waves using FlexSA, ISA support is necessary. First, an instruction to define the configurations of each FlexSA modes is needed. This instruction provides the type of FlexSA mode and the sizes of each wave's \(blk_M\), \(blk_N\), and \(blk_K\). This information is mapped to the micro-controls required to execute these waves such as data path selection, input shifting, output shifting, and accumulation. Second, vector load and store instructions are needed for efficient data transfers between the GBUF (global buffer) and LBUFs (local buffers). Using vectors reduces instruction count and instruction decoding BW, compared to using word-granularity instructions. The compiler generates memory access routines that load inputs to the LBUFs of cores and store the GEMM outputs to the GBUF or off-chip memory using these vector instructions.

\begin{algorithm}[!t]
\caption{GEMM execution flow}
\label{algo:complier}
\begin{algorithmic}[1]

\LineComment{\(blk_M\), \(blk_N\), \(blk_K\): GEMM tiling factors}
\For {\(n \gets 0\) to \(N\) by \(blk_N\)}
    \State \(n_{size}\) = \(blk_N\) {\bf{if}} \((n+1) \times blk_N < N\) {\bf{else}} \(N\) mod \(blk_N\)
    \For {\(m \gets 0\) to \(M\) by \(blk_M\)}
        \State \(m_{size}\) = \(blk_M\) {\bf{if}} \((m+1) \times blk_M < M\) {\bf{else}} \(M\) mod \(blk_M\)
        \State \(wide_{wave}\) = {\bf{IsWideWave}}(\(n_{size},\ m_{size}\))
        \For {\(k \gets 0\) to \(K\) by \(blk_K\)}
            \State \(k_{size}\) = \(blk_K\) {\bf{if}} \((k+1) \times blk_K < K\) {\bf{else}} \(K\) mod \(blk_K\)
            \State \(tall_{wave}\) = {\bf{IsTallWave}}(\(k_{size}\))
            \LineComment{Select FlexSA mode to execute the current wave}
            \State \(Mode\) = {\bf{GetFlexSAMode}}(\(wide_{wave},\ tall_{wave}\))
            \LineComment{Load stationary inputs to local buffers}
            \State {\bf{LdLBUF\_V}}(\(GBUF_{ptr1},\ LBUF_{ptr1},\ k_{size}, n_{size}\))
            \LineComment{Shift stationary inputs to each PE}
            \State {\bf{ShiftV}}(\(k_{size}, n_{size}\))
            \LineComment{Load inputs to horizontally shift to a local buffer}
            \State {\bf{LdLBUF\_H}}(\(GBUF_{ptr2},\ LBUF_{ptr2}, k_{size}, m_{size}\))
            \LineComment{Execute a wave using systolic dataflow}
            \State {\bf{ExecGEMM}}(\(Mmode,\ m_{size},\ n_{size},\ k_{size}\))
            \State {\bf{\twound{sync()}}}            
        \EndFor
    \LineComment{Store GEMM outputs to a memory}
    \State {\bf{StLBUF}}(\(OBUF_{ptr},\ GBUF_{ptr3}\))
    \EndFor
\EndFor

\end{algorithmic}
\end{algorithm}

\algo{algo:complier} shows how the FlexSA compiler processes a GEMM: generating instructions for data loading and FlexSA model selection. The compiler tiles the GEMM and schedules waves to process each GEMM tile. The type of each FlexSA mode is determined by the shape of the current systolic wave (\emph{IsWideWave} and \emph{IsTallWave}). 

Before executing each wave (or waves in parallel), inputs for the wave(s) are loaded from the GBUF to LBUFs of FlexSA cores. For this, two vector load instructions ({\emph{LdLBUF\_V}} and {\emph{LdLBUF\_H}}) are used. These vector load instructions take address pointers for GBUF and LBUFs, and the size of the data to load. The LBUFs are double buffered to hide the access and transfer latency. Thus, the inputs to be used in the next systolic wave are pre-fetched while executing the current wave. Once stationary inputs are fully loaded at LBUFs, they are shifted to PEs using a {\emph{ShiftV}} instruction. ShiftV takes a number of shifts and this is determined by the height of a wave. Using a separate instruction for stationary input shifting decouples it from the main systolic wave execution step and makes it parallel to loading non-stationary inputs to the LBUFs, removing unnecessary execution step serialization within a wave. 

When both inputs are ready, an {\emph{ExecGEMM}} instruction executes the target systolic wave using the selected FlexSA mode. ExecGEMM takes four inputs: the type of FlexSA mode and the sizes of the wave dimensions (\(blk_M\), \(blk_N\), and \(blk_K\)). For HSW, VSW, and ISW, the sizes of these wave dimensions indicate those of each independent wave. When GEMM tile execution is complete after iterating over the \(K\) dimension (processing all waves of a GEMM tile), the outputs accumulated at the OBUFs (output buffers) are stored to the GBUF using a {\emph{StLBUF}} instruction.



\section{Evaluation Methodology and \\Baseline Scheduling Policy}
\label{sec:flexsa_evaluation_method}

We develop an instruction-level architecture simulator to evaluate the performance and energy efficiency of FlexSA. We use three popular modern CNN models for evaluation: ResNet50, Inception v4~\cite{szegedy2017inception}, and MobileNet v2~\cite{sandler2018mobilenetv2}. ResNet50 is pruned during training using PruneTrain as the pruning mechanism with two different model-pruning strengths. Inception v4 is artificially pruned by applying the same pruning statistics of ResNet50. In case of MobileNet v2, we compare the training performance of the baseline model and its statically pruned version, which uses 75\% of channels in all convolution layers as used in the original proposal~\cite{sandler2018mobilenetv2}. The model is trained for 90 epochs with a pruning interval of 10 epochs. We use a mini-batch size of 32 for ResNet50 and Inception v4 and a larger mini-batch of 128 for MobileNet v2 considering their off-chip memory capacity requirements. We use mixed-precision multiplication and accumulation~\cite{micikevicius2017mixed} to execute the GEMMs of the convolution and FC layers.

In our evaluation, five different accelerator configurations are used, as summarized in \tab{tab:flexsa_config}. The {\bf{1G1C}} configuration uses a single 128$\times$128 systolic array, {\bf{1G4G}} splits this large core into four 64$\times$64 cores, which share a single GBUF. The {\bf{4G4C}} configuration further splits the cores into 16 32$\times$32 cores, where we divide the cores and the GBUF into four groups with cores in each group sharing a GBUF. The {\bf{1G1F}} configuration has a single 128$\times$128 FlexSA with four 64$\times$64 cores, and {\bf{4G1F}} has four small FlexSAs, each with a dedicated GBUF. The systolic array cores operate with 0.7 Ghz clock frequency so all configurations have 23 TFLOPS. We use a GBUF of 10 MB as used in WaveCore~\cite{lym2018mini} for all core configurations, and a single 270 GB/s HBM2 chip as the memory system~\cite{jedec2016hbm2}. All local input and output buffers are sized to support input and output double buffering. The local input buffers holding the horizontally shifted inputs are two times larger than the input buffers holding stationary inputs to support the larger reuse of the pre-loaded stationary inputs. 

\begin{table}[h]
\centering
\caption{Evaluation configuration description.}
    \noindent\resizebox{0.92\linewidth}{!}{
        \begin{tabu}{|[1.2pt]l|l|[1.2pt]}
            \thickhline
            \makecell[c]{Configuration} & \makecell[c]{Description} \tabularnewline
            \midhline
            1G1C                & 1 group each with 1$\times$ (128$\times$128) core \tabularnewline
            \hline
            1G4C                & 1 group each with 4$\times$ (64$\times$64) cores \tabularnewline
            \hline
            4G4C                & 4 groups each with 4$\times$ (32$\times$32) cores \tabularnewline
            \hline
            1G1F           & 1 group each with 4$\times$ (64$\times$64) FlexSA \tabularnewline
            \hline
            4G1F           & 4 groups each with 1$\times$ (32$\times$32) FlexSA \tabularnewline
            \thickhline
        \end{tabu}
    }
    \label{tab:flexsa_config}
\end{table}

DL models contain non-GEMM layers (e.g., feature normalization~\cite{ioffe2015batch,wu2018group}, activation~\cite{nair2010rectified}, and element-wise math layers), which tend to be bound by memory BW due to low arithmetic intensity. The execution time of these layers changes depending on accelerator design choices (off-chip memory BW and on-chip buffer design) and workload scheduling methods (multi-layer fusion~\cite{chen2018tvm}, and layer fusion~\cite{alwani2016fused,lym2018mini}). To decouple the impact of these memory bound operations and concentrate on what we intend to improve with FlexSA, we mainly focus on convolution and FC layers that use matrix operations and discuss the impact of other layers.

\medskip
\noindent\textbf{GEMM Partitioning and Blocking.}
The convolution and FC layers have three GEMM execution phases: forward propagation and data gradient and weight gradient computations in back-propagation. The GEMMs in the forward propagation and data gradient computation are skinny: they have large GEMM height (M), which is the product of the mini-batch size and the feature size, and have small GEMM width (N), which is set by the number of output channels. Thus, when multiple core groups are used, we partition a GEMM across the M dimension with one partition per core group. On the other hand, the GEMMs for weight gradient computation have small GEMM dimensions in both height and width, but they have large accumulation dimension (K). In this case, we partition the GEMM across the accumulation dimension with one partition per group. Within each GEMM partition, we use 2-level GEMM blocking that holds the inputs of a multiple of GMEM tiles in the GBUF for reuse. Within each GEMM partition, each systolic wave is allocated to cores in a group in round-robin fashion. When partitioning GEMM across the M dimension, different core groups use the same inputs from the GEMM N dimension. These shared inputs between core groups are replicated to avoid expensive inter-group data transfers. We find that this method effectively distributes GEMM workloads to multiple cores and core groups.

\section{Evaluation Results}
\label{sec:eval}

\noindent\textbf{PE Utilization and Performance.}
\fig{fig:flexsa_pe_util_ideal} shows the average PE utilization of ResNet50, Inception v4, MobileNet v2 across training epochs using different core configurations. This experiment uses infinite (ideal) DRAM BW to isolate the impact on PE utilization to only the size mismatch between systolic waves and cores. The baseline 1G1C shows low average ideal PE utilization of 44\% for the three CNNs. The PE utilization of Inception v4 and MobileNet v2 are lower than that of ResNet50 because many of their convolution layers have fewer than 128 channels and are smaller than the height and width of a core. Using one FlexSA (1G1F) improves the average PE utilization to 66\% and by 49\% compared to 1G1C because it processes the GEMM using tiles with higher PE occupancy so less internal PE utilization fragmentation. Using four FlexSA units (4G1F) further improves the PE utilization to 84\% and by 89\% compared to 1G1C showing diminishing return. Compared to 1G4C and 4G4C that use 4 and 16 small independent cores, 1G1F and 4G1F show only 0.1\% less ideal PE utilization. Considering that the PE utilization of using independent small cores is the maximum that can be achieved with FlexSA. The result indicates that the FlexSA modes selected by the proposed heuristics achieve near-optimal PE utilization.


\begin{figure*}[t!]
    \centering                                                                            
    \subfloat[Infinite (ideal) DRAM BW]{
        \includegraphics[width=0.495\textwidth]{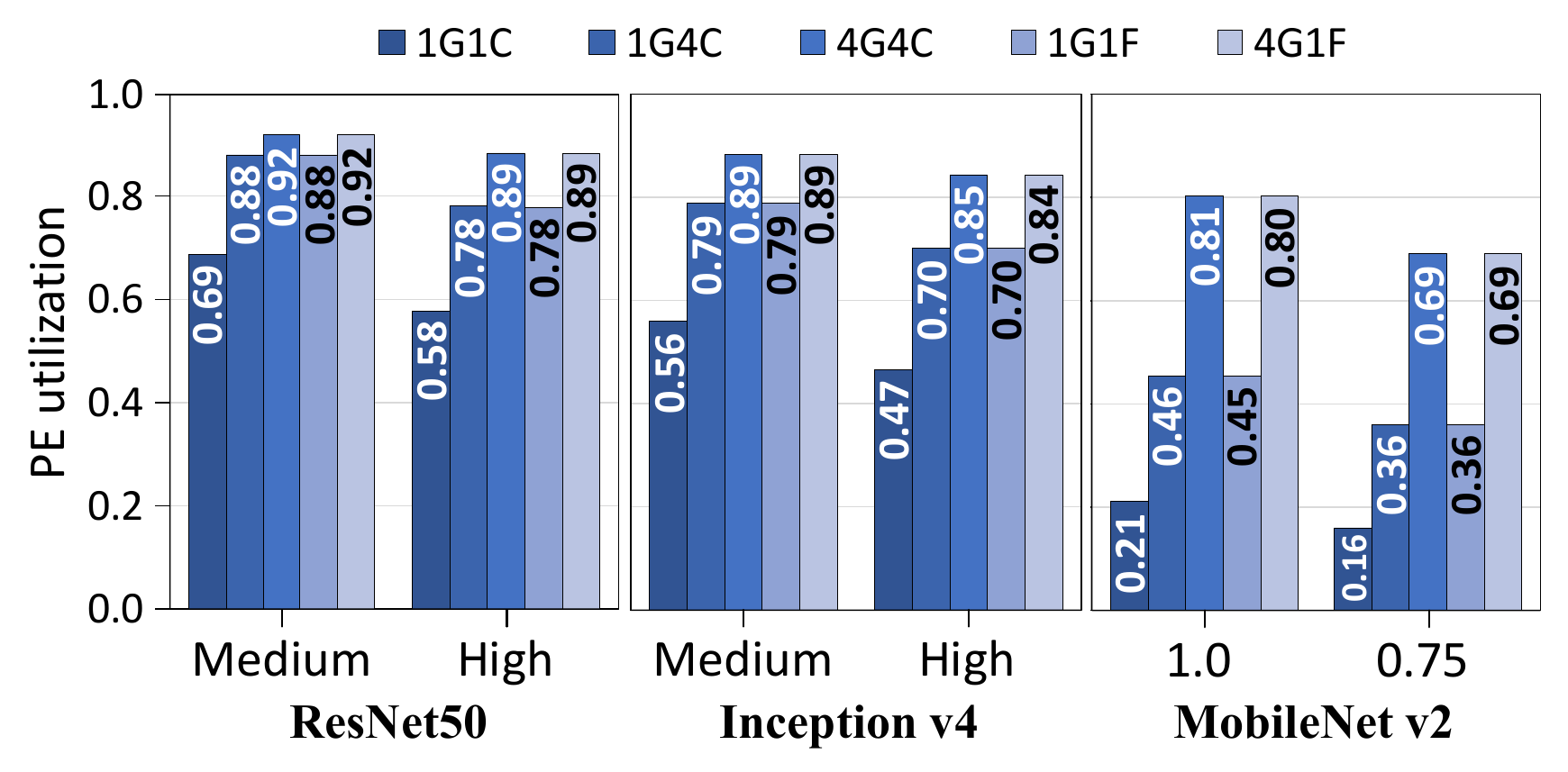}
        \label{fig:energy_breakdown}
    }
    \subfloat[Memory system with 1$\times$ HBM2]{
        \includegraphics[width=0.495\textwidth]{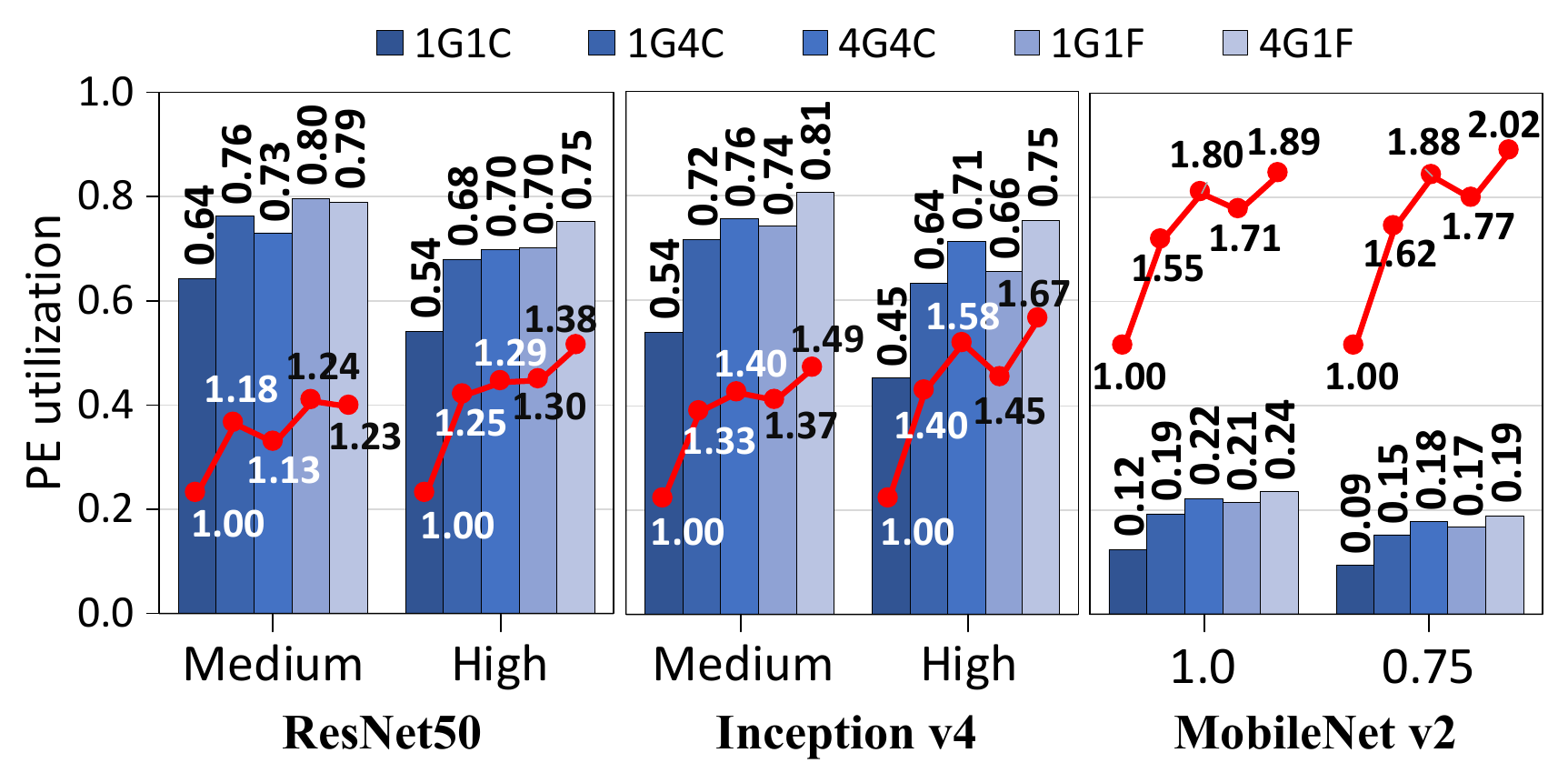}
        \label{fig:energy_breakdown2}
    }
    \caption{PE utilization of different core configurations and the speedup normalized to 1G1C (red lines).}
    \label{fig:flexsa_pe_util}
\end{figure*}

\fig{fig:flexsa_pe_util_hbm} shows the PE utilization and speedup using a single HBM2 for the DRAM system. Unlike using ideal memory BW, this reflects the performance impact from GEMM tiling, input blocking, and systolic wave scheduling. Therefore, for all configurations, PE utilization decreases compared to using ideal memory. Compared to using small independent cores (1G4C and 4G4C), both FlexSA configurations show smaller PE utilization drop contributing greater performance gains. This is because 1G4C and 4G4C always use small GEMM tiles and their performance suffers by increased DRAM BW peaks from executing many small independent systolic waves in parallel. FlexSA mitigates this overhead by using inter-core operating modes. Executing large waves using big cores increases arithmetic intensity per wave requiring low peak memory BW and small systolic waves are used only when using large waves decreases PE utilization. Overall, 1G1F and 4G1F achieve average pruning-while-training speedup of 37\% and 47\% compared to 1G1C and by 6\% and 7\% even compared to 1G4C and 4G4C, respectively. This is shown by red lines in \fig{fig:flexsa_pe_util_hbm}.

MobileNet v2 shows lower PE utilization compared to other CNNs because of its neural network architecture specifics. It uses pairs of depth-wise convolution layer and point-wise convolution layer, which have reduced input reuse. Thus, its performance becomes highly memory BW-bound with little on-chip reuse opportunity.

\begin{figure}[b!]
    \centering
    \includegraphics[width=0.48\textwidth]{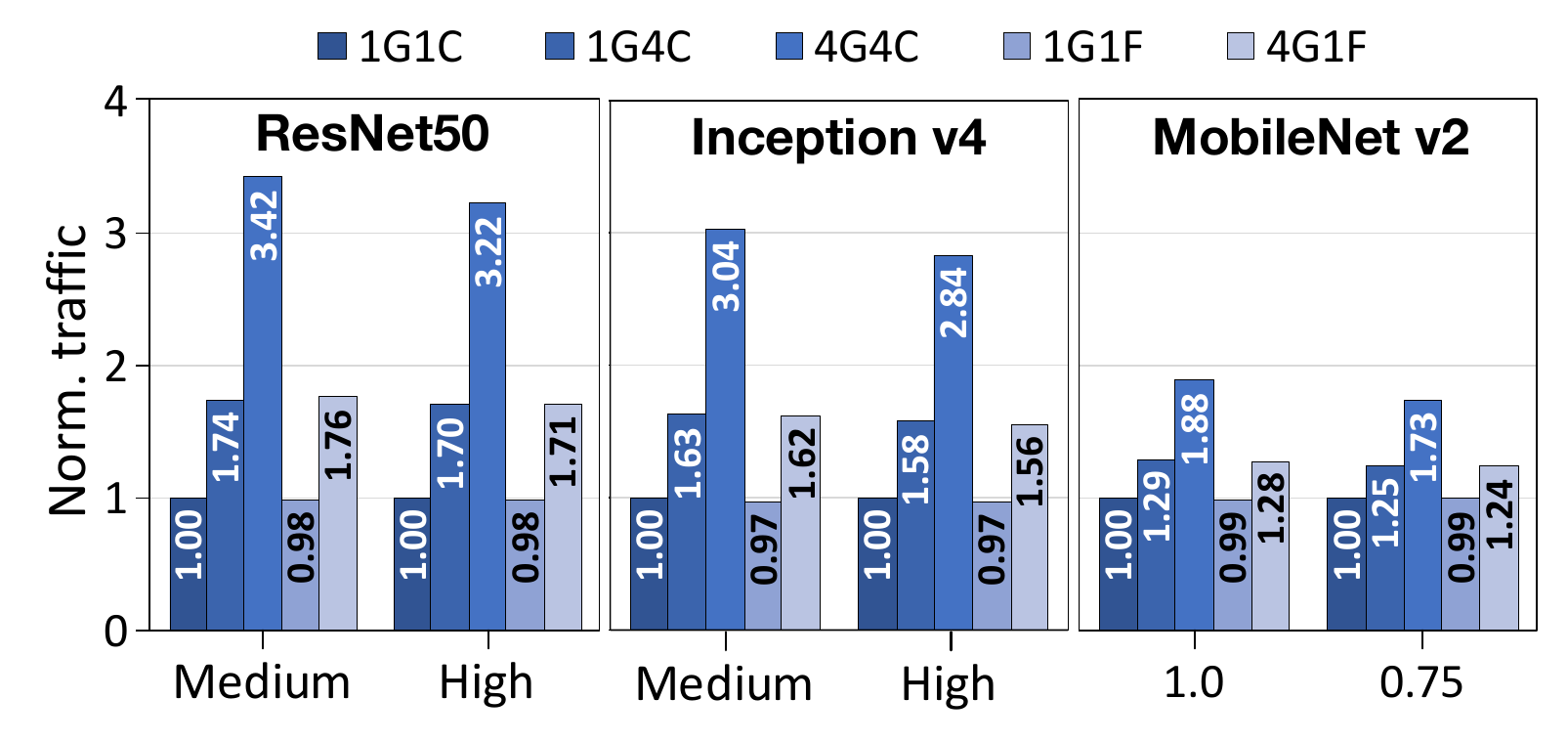}
    \caption[On-chip traffic comparison]{On-chip traffic of different core configurations and pruning strengths.}
    \label{fig:flexsa_gbuf_loads}
\end{figure}

\medskip
\noindent\textbf{On-chip Traffic and Energy Efficiency.}
Other than improving PE utilization, FlexSA reduces on-chip traffic by improving in-core reuse. \fig{fig:flexsa_gbuf_loads} compares the on-chip traffic between the GBUF to LBUFs of convolution and FC layers of in different models, which are normalized to that of 1G1C. The naive many-core configurations (1G4C and 4G4C) increase the data traffic by 1.5$\times$ and 2.7$\times$, respectively, due to the reduced in-core reuse caused by using small independent systolic waves. On the other hand, the configurations using FlexSA show far lower data traffic compared to the naive many-core designs and similar traffic even compared to the design with larger cores. 1G1F reduces on-chip traffic by 36\% compared to 1G4C, and by 2\% even compared to 1G1C. This 2\% traffic saving comes from reusing the stationary inputs and outputs between multiple systolic waves in processing VSW and ISW modes, which is not possible with the baseline core design. 4G1F also saves on average 43\% GBUF traffic compared to 4G4C.

The reduced on-chip traffic from using FlexSA improves the energy efficiency of pruning and training. To show this, we evaluate the dynamic energy consumed per training iteration using different core configurations (\fig{fig:energy_breakdown}). Each bar indicates the breakdown of energy consumed by different accelerator operations. \emph{COMP} indicates the energy consumed by mixed-precision multiply and accumulation operations of the FlexSA core. \emph{LBUF}, \emph{GBUF}, and \emph{DRAM} indicate the energy spent on data transmission associated with different resources. \emph{OverCore} indicates the over-core data transmission energy, which exists only in the configurations using FlexSA. The lines represent the energy increase compared to 1G1C. Both FlexSA configurations exhibit similar or even lower energy consumption than the baseline 1G1C. This high energy efficiency of FlexSA comes from using large systolic waves in most cases. The additional energy consumed by over-core data transmission is very small for all CNNs and pruning strengths. However, 1G4C and 4G4C exhibit $>$20\% total energy increase for ResNet50 and Inception v4 compared to both FlexSA configurations due to reduced in-core reuse and replicated input loads between cores. The reason 4G4C shows similar energy compared to 1G4C though it requires higher traffic is that average global buffer access energy is lower with the distributed GBUFs. However, MobileNet v2 shows smaller energy increase with 1G4C and 4G4C shows even lower energy consumption than 1G1C. This is because DRAM access accounts for the dominant energy consumption of MobileNet v2 and GBUF energy takes relatively small portion with low on-chip reuse.


\begin{figure}[t!]
    \centering                                                                            
    \subfloat[Medium pruning strength]{
        \includegraphics[width=0.48\textwidth]{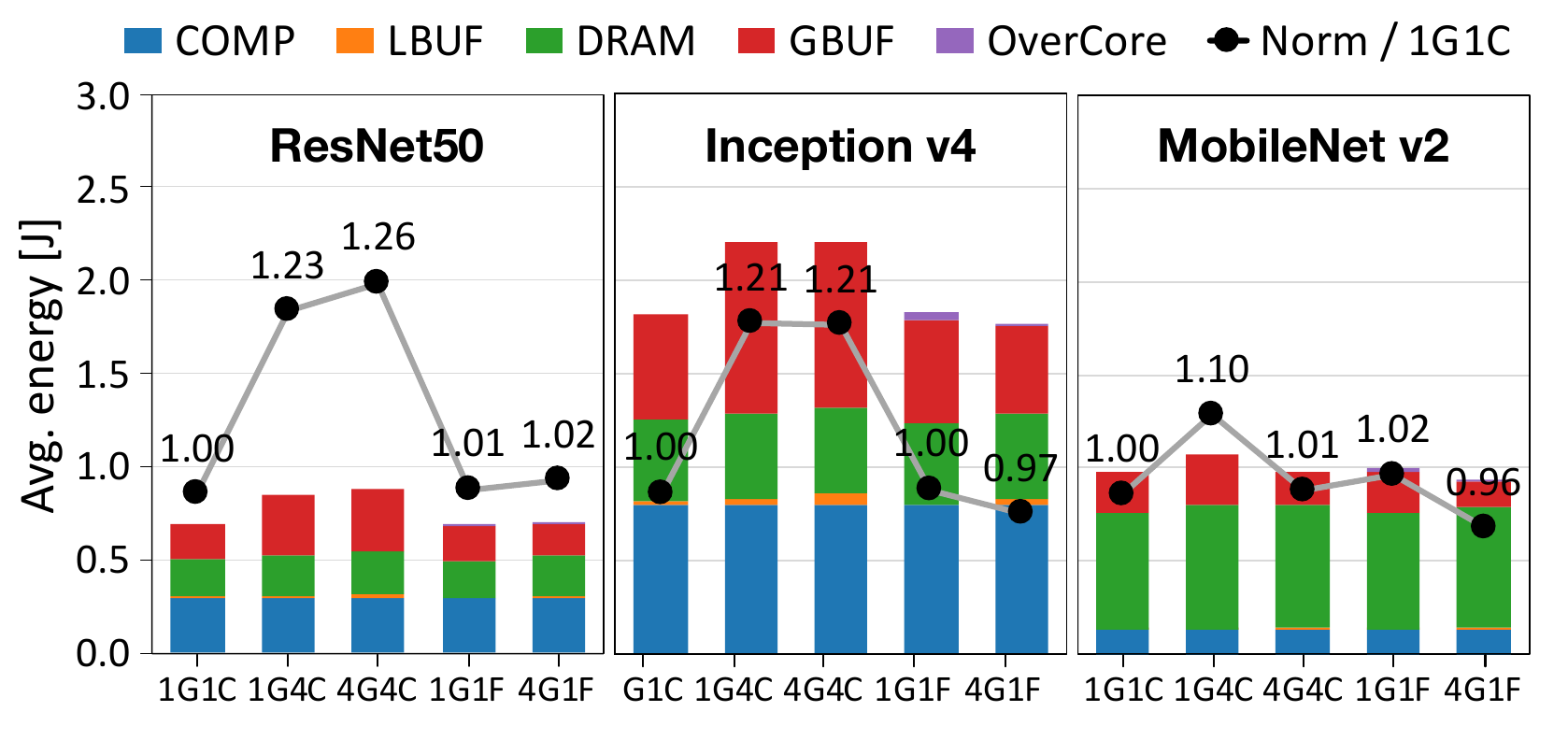}`
        \label{fig:energy_breakdown}
    }\\
    \vspace*{-2.0mm}
    \subfloat[High pruning strength]{
        \includegraphics[width=0.48\textwidth]{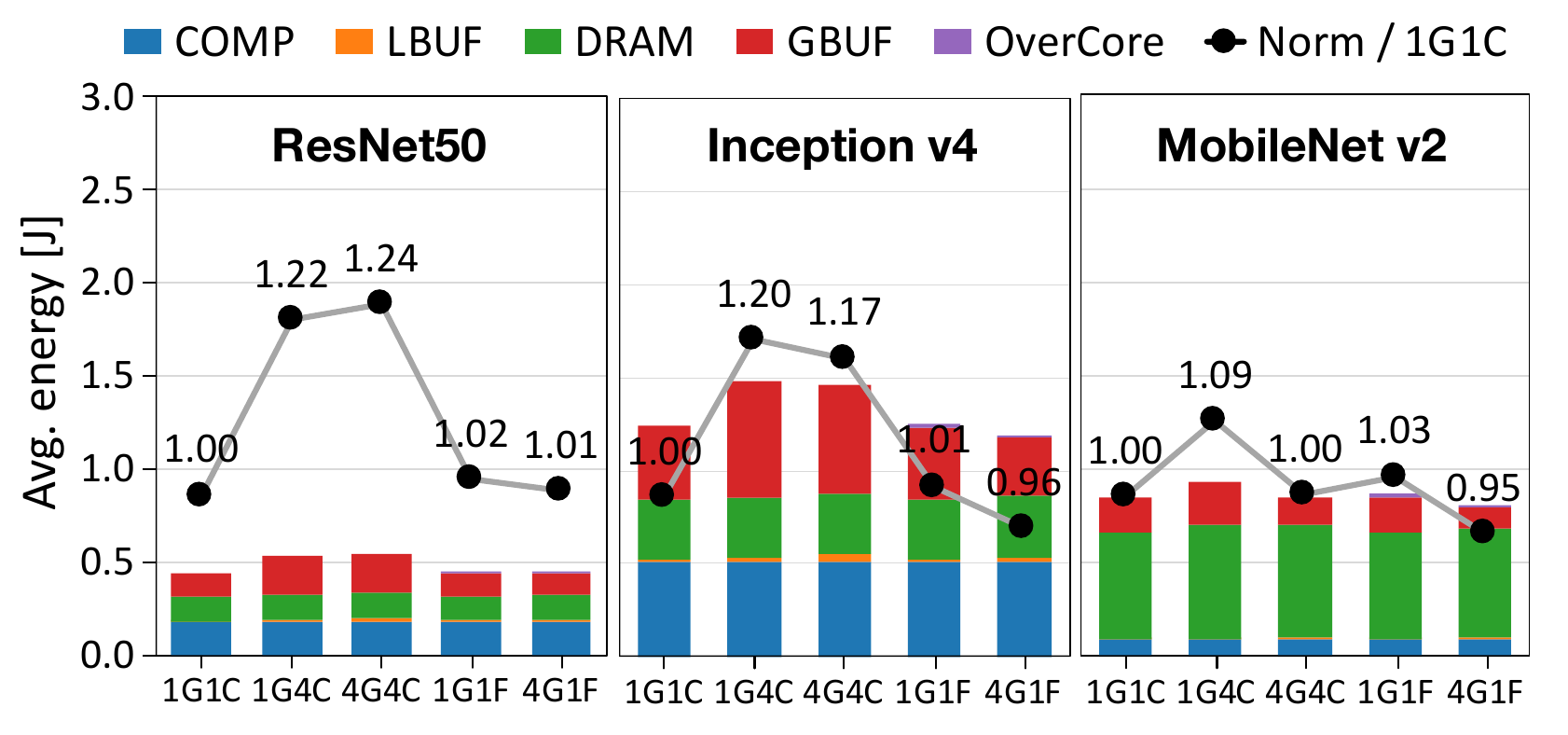}
        \label{fig:energy_breakdown2}
    }
    \caption{Breakdown of dynamic energy consumption per training iteration with different core configurations and pruning strength.}
    \label{fig:fig:flexsa_pe_util}
\end{figure}

\begin{figure}[t!]
    \centering
    \includegraphics[width=0.48\textwidth]{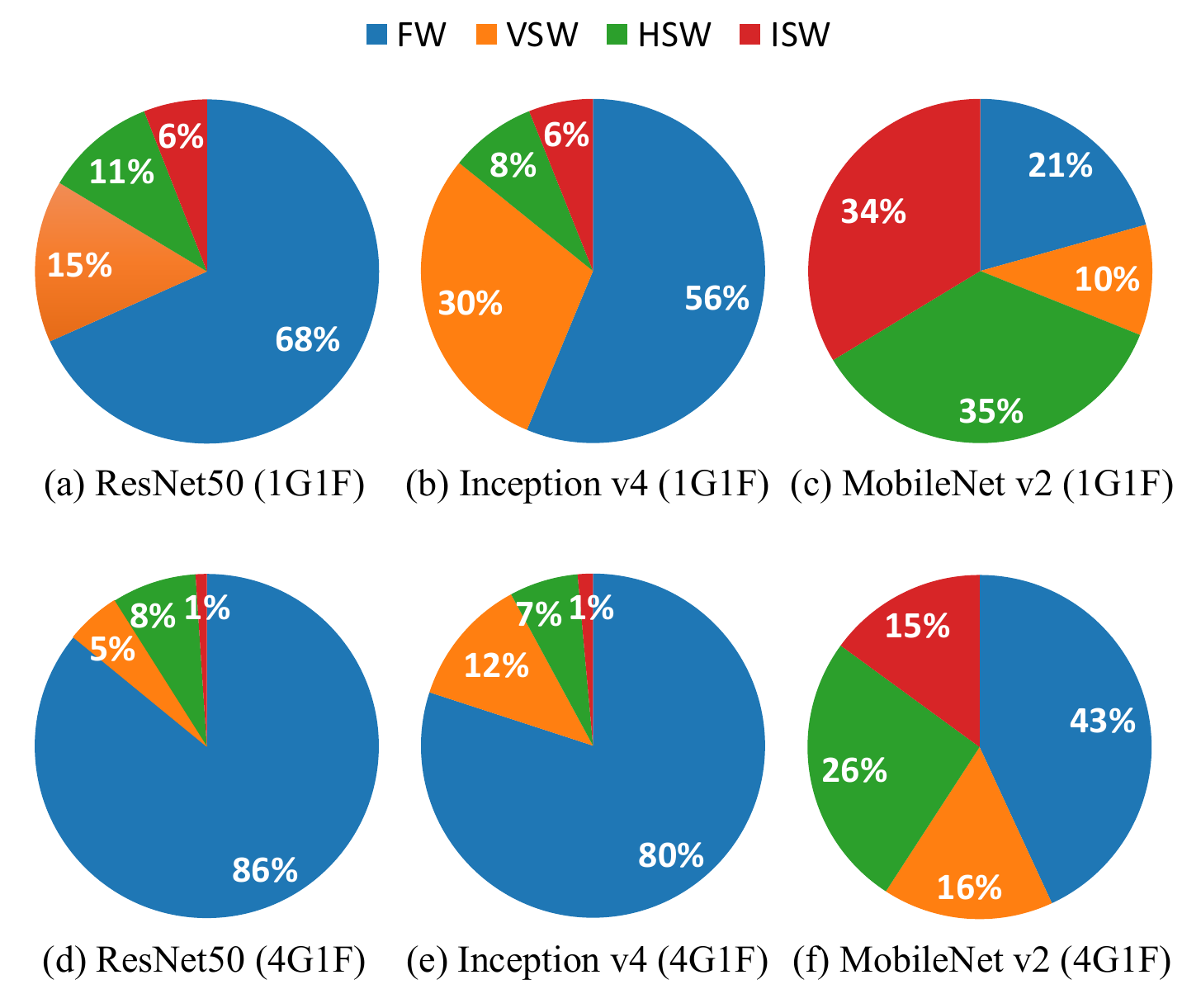}
    \caption[FlexSA modes selection breakdown]{FlexSA operating modes breakdown of pruning-while-training using 1G1F and 4G1F.}
    \label{fig:inst_breakdown}
\end{figure}

\medskip
\noindent\textbf{FlexSA Operating Modes Breakdown.}
\fig{fig:inst_breakdown} shows the breakdown of FlexSA operating modes used in pruning with 1G1F (top row) and 4G1F (bottom row) configurations. Each pie graph averages the breakdown of using high and low pruning strengths. This shows how frequently large systolic waves are used to improve in-core reuse and save energy consumption. For 1G1F, the inter-systolic array operating modes (VSW, HSW, and FW) account for 94\% for both ResNet50 and Inception v4 and 66\% for MobileNet v2. For 4G4F, the ratio of using inter-systolic array operating modes increase to 99\% for both ResNet50 and Inception v4 and to 85\% for MombileNet v2. Especially, for ResNet50 and Inception v4, the least efficient ISW accounts for only 6\% and 1\% for 1G1F and 4G1F. MobileNet v2 shows a bit greater ISW usage because it uses tensors with small dimensions. This result indicates that the high-reuse inter-core operations of FlexSA are used frequently and still achieve very high utilization.

\medskip
\noindent\textbf{Performance and Energy Impact of Other Layers.}
We evaluate the impact of other layers without layer fusion. This is a conservative assumption given many of memory-bound math layers can be executed while executing GEMMs. We use SIMD array with 500 GFLOPS to execute the data-parallel operations, which is the 1/50 of the throughput of the systolic arrays. Based on our evaluation, 1G1F and 4G1F improves end-to-end training performance of all three CNNs by 24\% and 29\%, respectively, compared to 1G1C. This is also 3\% performance improvement compared to the simple many-small-core designs. This performance gain will increase when aggressive layer fusion is considered~\cite{chen2018tvm}. The FlexSA configurations also save overall dynamic energy in training by 10\% compared to the many-small-core designs.

\section{Conclusion}
\label{sec:conclusion}

In this paper, we tackle the problems in training (partially) pruned CNN models using a high-throughput accelerator with systolic arrays. The GEMMs of structurally pruned CNN models have reduced dimensions, which decreases PE utilization when processed on large systolic arrays. Although splitting a large systolic array into multiple small arrays improves PE utilization, it decreases in-core data reuse, increasing on-chip traffic and energy consumption. 

To improve the PE utilization without affecting reuse, we present FlexSA, a flexible systolic array architecture that dynamically reconfigures its structure to efficiently process systolic waves with different sizes and shapes. FlexSA consists of four sub-systolic array cores and designing it adds only 1\% area compared to a naive four-core design. FlexSA supports four different systolic array operating modes and each is designed to efficiently process systolic waves with different shapes, forming different inter-core systolic dataflows. Our FlexSA compiler supports optimal utilization of the underlying FlexSA resources using an efficient mode selection heuristic and custom instructions. Overall, FlexSA improves PE utilization of pruning while training CNN models by 37\% compared to using a single large core and increase on-chip reuse by 1.7X compared to small-many-core designs, saving dynamic energy by 28\%.



\bibliographystyle{IEEEtran}
\bibliography{bib/ref}
\end{document}